\documentclass[letterpaper]{article} 
\usepackage{aaai25} 
\usepackage{times} 
\usepackage{helvet} 
\usepackage{courier} 
\usepackage[hyphens]{url} 
\usepackage{graphicx} 
\usepackage{natbib} 
\usepackage{caption} 
\usepackage{algorithm}
\usepackage{algorithmic}

\usepackage{newfloat}
\usepackage{listings}
\DeclareCaptionStyle{ruled}{labelfont=normalfont,labelsep=colon,strut=off} 
\floatstyle{ruled}
\newfloat{listing}{tb}{lst}{}
\floatname{listing}{Listing}
\usepackage[table,xcdraw, dvipsnames]{xcolor}
\usepackage{multirow}
\usepackage{diagbox}
\usepackage{amssymb}
\usepackage{amsmath}
\usepackage{lipsum} 
\usepackage{tcolorbox}
\tcbuselibrary{theorems}
\tcbuselibrary{breakable}
\usepackage{booktabs}
\usepackage{listings}
\definecolor{dark_red}{cmyk}{0, 1.0, 1.0, 0}
\definecolor{light_red}{cmyk}{0, 0.75, 0.75, 0}
\definecolor{dark_green}{cmyk}{1.0, 0.0, 1.0, 0.5}
\definecolor{light_green}{cmyk}{0.7, 0.0, 0.7, 0.45}
\definecolor{dark_blue}{cmyk}{1.0, 1.0, 0.0, 0.0}
\definecolor{light_blue}{cmyk}{0.85, 0.85, 0.0, 0.0}
\definecolor{dark_mag}{cmyk}{0.0, 1.0, 0.0, 0.25}
\definecolor{light_mag}{cmyk}{0.0, 0.75, 0.0, 0.25}

\title{Are Large Language Models Strategic Decision Makers? \\ A Study of Performance and Bias in Two-Player Non-Zero-Sum Games}

\author{
  Nathan~Herr\textsuperscript{\rm 1},
  Fernando~Acero\textsuperscript{\rm 1},
  Roberta~Raileanu\textsuperscript{\rm 1,2},
  María~Pérez-Ortiz\textsuperscript{\rm 1},
  Zhibin~Li\textsuperscript{\rm 1}
}
\affiliations{
\small{\textsuperscript{1}University College London}, \textsuperscript{2}FAIR at Meta
 \\
 \small{
   \texttt{ \{nathan.herr.19, fernando.acero, maria.perez, alex.li\}@ucl.ac.uk}, \texttt{raileanu.roberta@gmail.com}
 }
}

\usepackage{bibentry}

\begin{document}
\maketitle

\begin{abstract}
Large Language Models (LLMs) have been increasingly used in real-world settings, yet their strategic decision-making abilities remain largely unexplored. To fully benefit from the potential of LLMs, it's essential to understand their ability to function in complex social scenarios. Game theory, which is already used to understand real-world interactions, provides a good framework for assessing these abilities. This work investigates the performance and merits of LLMs in canonical game-theoretic two-player non-zero-sum games, Stag Hunt and Prisoner Dilemma. Our structured evaluation of GPT-3.5, GPT-4-Turbo, GPT-4o, and Llama-3-8B shows that these models, when making decisions in these games, are affected by at least one of the following systematic biases: positional bias, payoff bias, or behavioural bias. This indicates that LLMs do not fully rely on logical reasoning when making these strategic decisions. As a result, it was found that the LLMs' performance drops when the game configuration is misaligned with the affecting biases. When misaligned, GPT-3.5, GPT-4-Turbo, GPT-4o, and Llama-3-8B show an average performance drop of 32\%, 25\%, 34\%, and 29\% respectively in Stag Hunt, and 28\%, 16\%, 34\%, and 24\% respectively in Prisoner's Dilemma. Surprisingly, GPT-4o (a top-performing LLM across standard benchmarks) suffers the most substantial performance drop, suggesting that newer models are not addressing these issues. Interestingly, we found that a commonly used method of improving the reasoning capabilities of LLMs, chain-of-thought (CoT) prompting, reduces the biases in GPT-3.5, GPT-4o, and Llama-3-8B but increases the effect of the bias in GPT-4-Turbo, indicating that CoT alone cannot fully serve as a robust solution to this problem. We perform several additional experiments, which provide further insight into these observed behaviours.
\end{abstract}

\section{Introduction}

\begin{figure}[ht]
\begin{center}
\centerline{\includegraphics[width=\columnwidth]{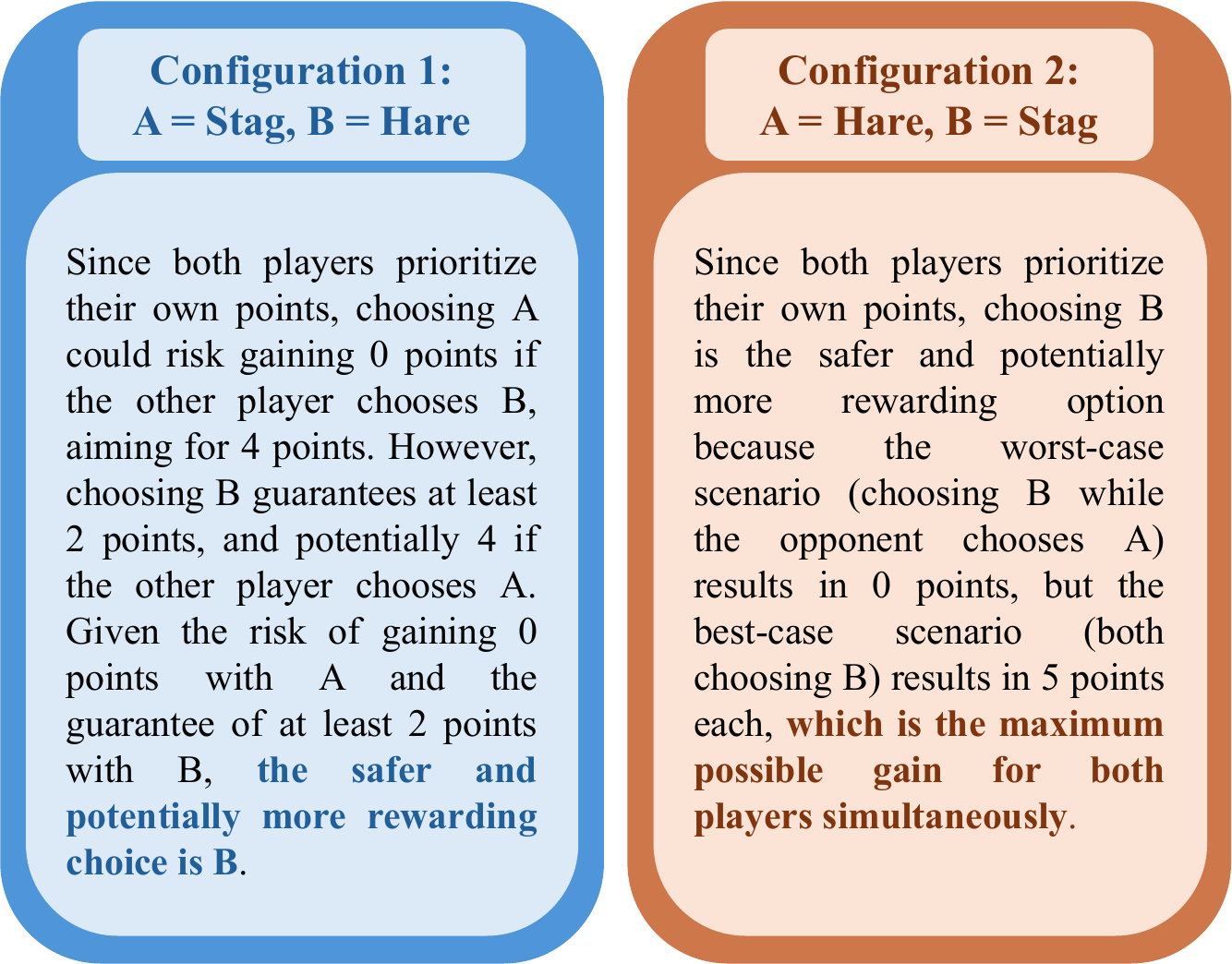}}
\caption*{
An LLM is prompted to select the best action in both \textsc{\textbf{Configuration 1}} and \textsc{\textbf{Configuration 2}}, where the order of the actions is reversed. 
Despite the same underlying task in both configurations, the LLM's reasoning differs, revealing that the model relies more on the order of the actions rather than their intrinsic meaning during the decision-making process. More examples can be found in Section~\ref{ssec:app_res_tabs} of the Technical Appendix.}
\label{icml-historical}
\end{center}
\end{figure}

The use of Large Language Models (LLMs) to solve everyday tasks has become increasingly widespread \cite{naveed2023comprehensive, hadi2023survey, duan2024gtbench}. As LLMs are deployed in the real world, interacting with other humans or artificial agents, there is a need to understand their capabilities to operate in complex social scenarios. To this end, this study chose game theory, which is already applied to many existing real-world tasks \cite{martin2017course} in fields such as economics, psychology, and sociology \cite{fan2024can, duan2024gtbench}, to investigate the decision-making capabilities of LLMs. 

Game theory \cite{roughgarden2010algorithmic} abstracts complex real-life scenarios as mathematical models designed to be easy to understand but require strategic decision-making abilities to be mastered. A sub-genre of game-theoretic games is two-player non-zero-sum games which have both competitive and cooperative elements. This provides a fair representation of agent interactions for many important real-world scenarios. LLMs have been shown to reach levels of reasoning comparable to humans, demonstrating their ability to solve such real-world tasks \cite{guo2024large}. These accomplishments are primarily achieved using prompts that are specifically curated for the task at hand, referred to as prompt engineering \cite{sahoo2024systematic}. 

The majority of techniques make use of prompt engineering, which has shown promise but is often not robust \cite{guo2024large}. This brittleness has been discovered and investigated in many state-of-the-art (SOTA) LLMs \cite{chen2024premise, Papadatos_Freedman, zheng2023large, zheng2024judging, wang2023large} such as LLama-2, GPT-3.5, and GPT-4, primarily in multiple-choice question answering and similar tasks. Specifically, these works show that LLMs suffer from what we refer to as systematic biases, which guide the LLMs' decisions and not their strategic reasoning. Far fewer studies have examined systematic biases within the context of game theory, an important field where such investigations are necessary to fully benefit from the use of LLMs in real-world tasks.

To date, there is a lack of research on the systematic biases of LLMs when solving game-theoretic tasks, which has motivated this work. Our objectives are to (1) investigate the systematic biases in several SOTA LLMs during two-player zero-sum games and analyze how these biases vary between models, (2) demonstrate the statistical evaluations of these biases within each model and assess their impact on performance, and (3) conduct further study to investigate and understand these behaviours, including: (i) exploring the impact of fine-tuning on these biases, (ii) testing how prior knowledge affects LLM biases, (iii) evaluating how different prompt structures influence observed biases, and (iv) comparing the behaviours of humans and LLMs in a two-player zero-sum game. 

\section{Related Work}

Having established the importance of investigating how systematic biases affect LLMs' ability to make strategic decisions in two-player non-zero-sum games, we now explore existing related work.

\paragraph{\textsc{LLMs and Game Theory.}} In recent years, the use of LLMs as single-agent planners/decision makers has evolved into LLM-based multi-agent systems \cite{guo2024large} where agents are required to solve strategic and logical reasoning problems. These capabilities are often evaluated through game-theoretic tasks \cite{zhang2022automatic, lore2023strategic, gandhi2023strategic}. This shift has prompted many new benchmarks testing LLMs in game theoretic tasks with the aim of progressing the work within the field \cite{xu2023magic, huang2024far, chen2023put, duan2024gtbench, li2023static, aher2023using}. Furthermore, several existing works focus on game theoretic matrix games, such as Prisoner's Dilemma, Stag Hunt, and Dictator Game, to name a few \cite{fan2024can, xu2023magic, lore2023strategic, brookins2023playing, gandhi2023strategic, phelps2023investigating, guo2023gpt}, which are discussed below. In \citet{fan2024can}'s work they show that LLMs, even when explicitly given the correct belief, from which they should reason to take correct action, tend to ignore or modify this belief. They also note that LLMs tend to select specific action labels more frequently than others (they note that GPT-3 prefers U to V), but do not investigate this any further. \citet{xu2023magic} and \citet{brookins2023playing} show that LLMs tend to select the cooperative action more frequently than humans, despite it not being the optimal choice in most cases. Lastly, several works test how the LLMs' behaviour changes as they modify the LLMs' preferences or contextual frameworks, such as being selfish or cooperative \cite{fan2024can, phelps2023investigating, guo2023gpt, lore2023strategic}. They all note that LLMs are seemingly capable of following simple preferences, such as selecting the selfish action when prompted to be selfish. However, they do not investigate the effect the chosen prompt configuration has on the LLMs' performance.

\paragraph{\textsc{Bias in LLMs.}} The presence of systemic biases (the term ``bias'' denotes the consistent deviation within LLMs \cite{zheng2023large}, such as token bias, which favours a specific token U over token V \cite{fan2024can}) has recently become a topic of interest. Specifically, these biases are found and tested in multiple choice question evaluation \cite{zheng2023large}, multi-turn question answer evaluation \cite{zheng2024judging}, response quality evaluation \cite{wang2023large}, and tasks such as text classification, fact retrieval, and information extraction \cite{zhao2021calibrate, chen2024premise, berglund2023reversal, golovneva2024reverse}. It was found that LLMs suffer from what is referred to as selection bias \cite{zheng2023large, zheng2024judging, wang2023large, zhao2021calibrate}, which is a combination of both token bias and positional bias. Prior works have also studied other types of biases when using LLMs to judge the quality of LLM generations. For example, \citet{zheng2024judging} discovers LLMs have a verbosity bias, favouring longer responses over shorter ones. Similarly, \citet{zheng2024judging} found that LLMs have a self-enhancement bias, favouring responses generated by the judge LLM relative to other LLMs. These are less relevant to our setting since we don't focus on LLMs that judge other LLMs' outputs. 



\section{Methodology}
\label{sec:meth}

\subsection{Two-Player Non-Zero-Sum games}
\label{ssec:games}

Prisoner's Dilemma is one of the most popular two-player non-zero-sum games used by researchers in both scientific literature and real-world settings to study complex social interactions \cite{gilbert1996prisoner, pacheco2009evolutionary, mantas2022factors}. Stag Hunt is frequently compared to Prisoner's Dilemma \cite{pacheco2009evolutionary} because of its similar social dynamics and the differences in their Nash Equilibria \cite{chatterjee2004two}. This work, therefore, focuses on these canonical two-player non-zero-sum games. A brief description of each game is provided below with a detailed explanation of the games in Section~\ref{ssec:app_game} of the Technical Appendix.

\paragraph{\textsc{Stag Hunt.}} A scenario involving two hunters who can either hunt a stag together or each hunt a hare individually \cite{skyrms2001stag}. They must decide their actions without communicating, based on their expectations of whether the other hunter will cooperate to hunt the stag or act independently to hunt the hare.

\paragraph{\textsc{Prisoner's Dilemma.}} A scenario where two prisoners must choose to confess or remain silent \cite{tucker1983mathematics}. If both confess, they each get 2 years in prison. If both remain silent, they each get 1 year. If one confesses and the other remains silent, the confessor goes free, and the silent one gets 3 years.

\paragraph{\textsc{Game and Player Descriptions.}} Both games are described using the same prompt skeleton (detailed in Section~\ref{ssec:app_prompt} of the Technical Appendix). The prompt makes use of action labels A and B to describe the payoff matrix for the specific games. For example, in Prisoner's Dilemma, the text \textit{``if you select action label A and they select action label B, you receive 0 points while they receive 3 points''} describes a scenario where you, the Acting Player (AP), selects to cooperate while they, the Fellow Player (FP), selects to defect. Note that neither the terms, cooperate nor defect are used in the prompt. The reason for this is to promote reasoning over the payoff matrix and not rely on the LLMs' prior knowledge of the games. To this effect, the names of each game are not explicitly mentioned in the prompt and are only identifiable by their payoff matrices. In addition to the description of the game, we include \textit{``player preferences''}, where the AP is prompted with the preferred behaviours (players either prioritise the common-gain or their self-gain) of both players, such that we may test how the LLMs' behaviour changes as we modify the preferred behaviours, much like the work done in \cite{fan2024can, phelps2023investigating, guo2023gpt, lore2023strategic}.

\subsection{LLM Biases}
\label{ssec:bias}

There are 3 ways in which one can change the prompt, used to describe the games, without changing the skeleton itself: (1) the order in which the action labels are presented in the prompt, (2) the payoffs associated with the action labels, and (3) the prompted preferred behaviours of the players. Given this, we focus our investigation on the biases related to these changes. The details of which are discussed below.

\paragraph{\textsc{Positional Bias.}} Changing the order in which the action labels are stated in the prompt affects the frequency of the selected action label. For example, Llama-3-8B, when prompted with action label A first and B second, tends to select the first action label A more frequently. 

\paragraph{\textsc{Payoff Bias.}} Changing the payoffs associated with the different action labels, PayOff(label\_1, label\_2), affects the frequency of the selected action label. In particular, a model may be biased towards, (1) selecting the action that leads to the maximum possible self-gain or (2) selecting the action that leads to the maximum possible common-gain, rather than the action that maximizes the expected gain (which takes into account all possible actions the other agent can take). Note that in Stag Hunt these two actions are the same and for Prisoner's Dilemma they are not (details on this can be found in Table \ref{tab:contingency_app} in Section~\ref{ssec:app_game} of the Technical Appendix). For example, GPT-4-Turbo in Prisoner's Dilemma, when PayOff(A, A)=2, PayOff(A, B)=0, PayOff(B, A)=3, and PayOff(B, B)=1, tends to select action label A (the action associated with the maximum possible common-gain). Another example, Llama-3-8B in Prisoner's Dilemma, with the same payoffs, tends to select action label B (the action associated with the maximum possible self-gain). 

\paragraph{\textsc{Behavioural Bias.}} Changing the preferred behaviour of the Acting Player (AP) and Fellow Player (FP) affects the frequency of the selected action. For example, when GPT-3.5 (AP) is prompted to prioritise Common-Gain (CG) it tends to select action label A, irrespective of the FP's preferred behaviour, and when prompted to prioritise Self-Gain (SG), tends to select action label B.

\begin{figure*}[ht!]
\begin{center}
\includegraphics[width=0.45\textwidth]{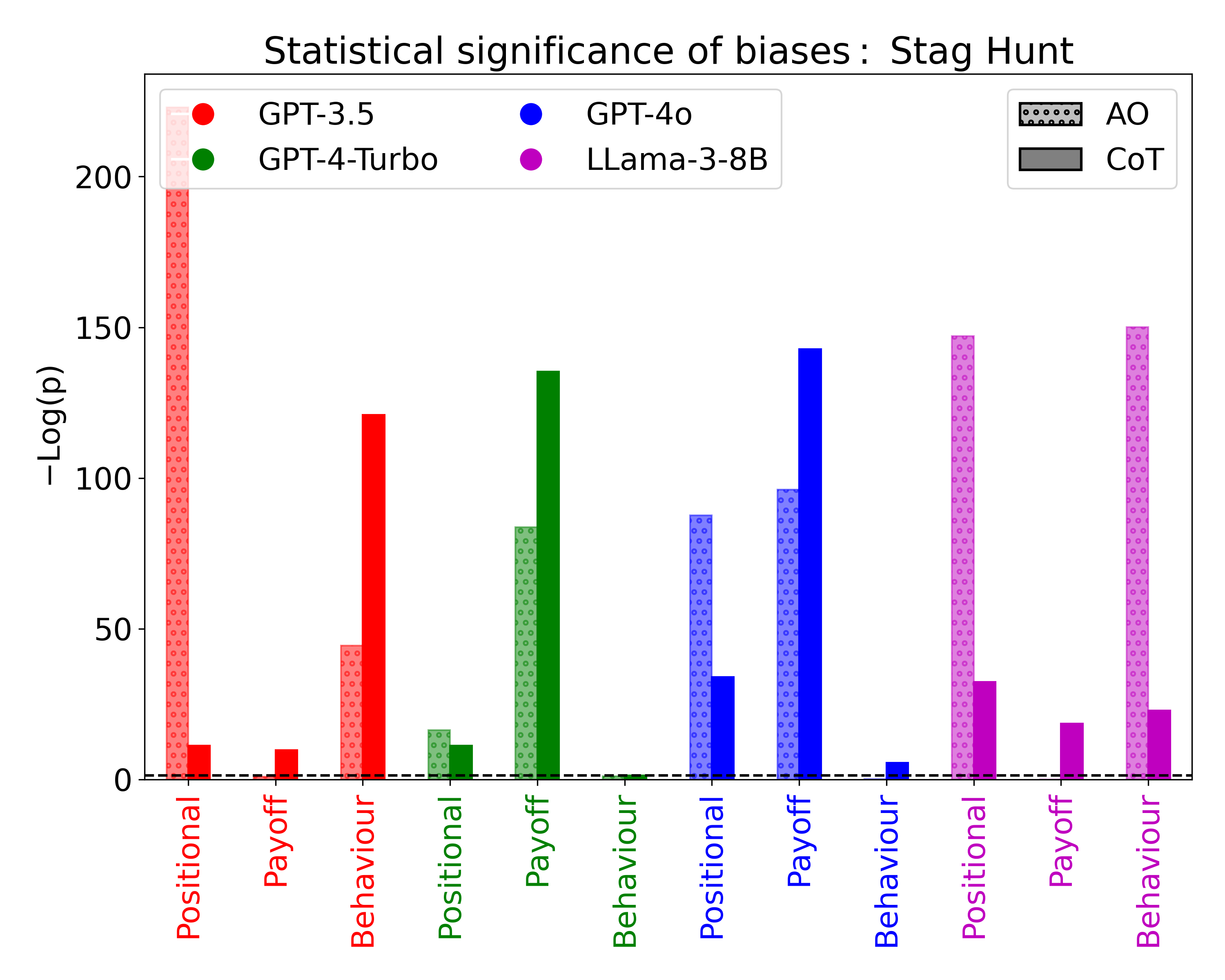}
\includegraphics[width=0.45\textwidth]{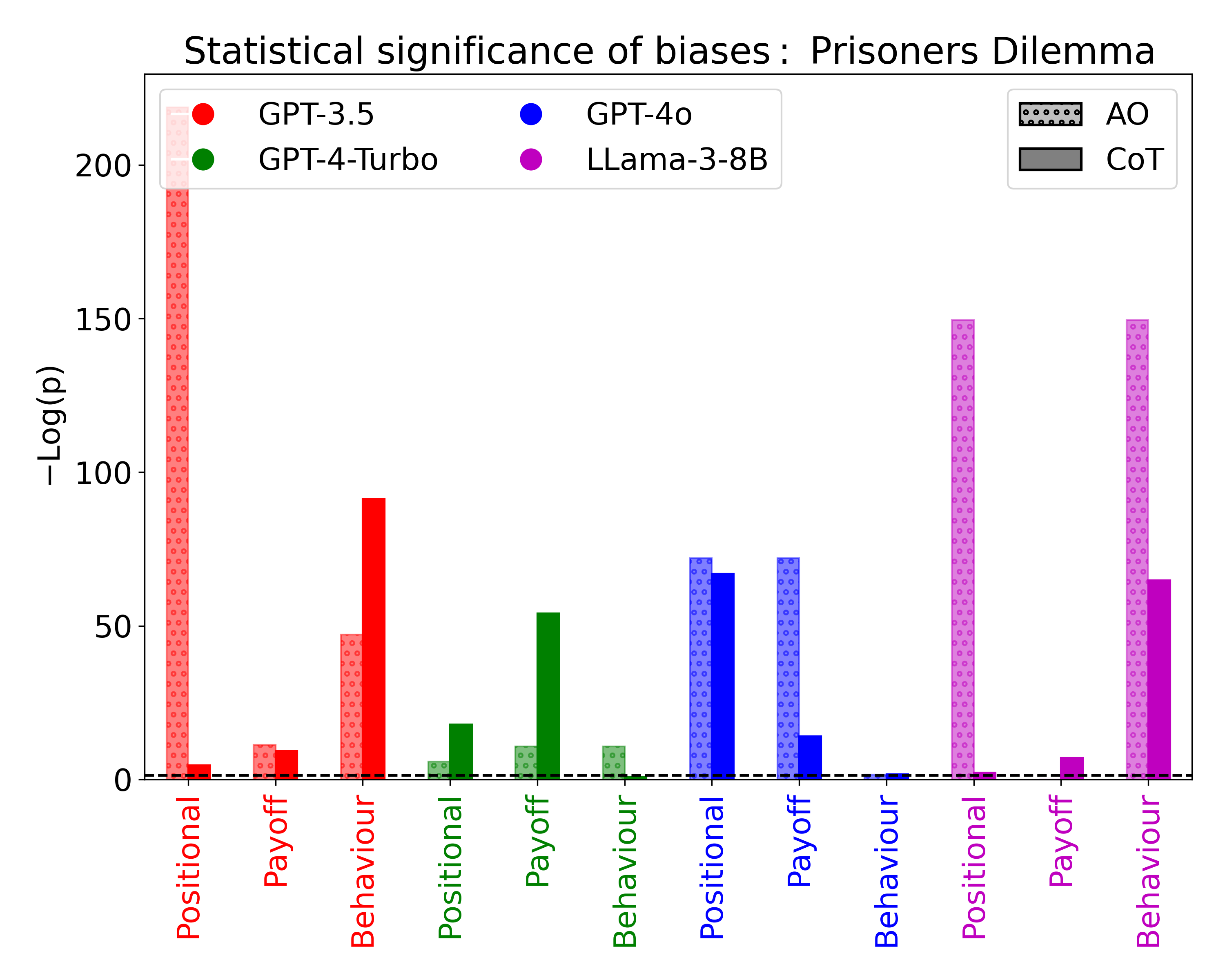}
\caption {Figure showing the statistical analysis of the identified biases for all models tested, \textbf{{GPT-3.5}}, \textbf{{GPT-4-Turbo}}, \textbf{{GPT-4o}}, and \textbf{{Llama-3-8B}}. The larger the $-Log(p)$, the more statistically significant the bias. The dashed \textbf{black} line signifies the threshold at which the bias becomes statistically significant (found close to the horizontal axis for both plots). Notably, it can be seen that each model is significantly affected by at least one of the identified biases under both prompting methods. We can also consider the average over all three biases for each prompting method, namely; (LEFT) Stag Hunt - \textbf{AO:} \textbf{{89.4}}, \textbf{{33.7}}, \textbf{{61.3}}, \textbf{{99.0}} and \textbf{CoT:} \textbf{{47.3}}, \textbf{{49.3}}, \textbf{{60.8}}, \textbf{{24.6}} and (RIGHT) Prisoner's Dilemma - \textbf{AO:} \textbf{{92.4}}, \textbf{{9.11}}, \textbf{{48.6}}, \textbf{{99.7}} and \textbf{CoT:} \textbf{{35.0}}, \textbf{{24.3}}, \textbf{{27.5}}, \textbf{{24.7}}. We can see that all models, except for GPT-4-Turbo, are affected less by the biases when using CoT prompting.}
\label{fig:stat_sig}
\end{center}
\end{figure*}

\subsection{Experimental Setup}

To perform this investigation, we methodically adjust the base prompt (detailed in Section~\ref{ssec:app_prompt} of the Technical Appendix) over all combinations of positions, payoffs, and behaviours, making up 16 different experimental setups. Additionally, each of these experimental setups is run with and without prompting the LLM to first reason over the problem. To do this, we use the following prompting schemes; (1) Answer-Only (AO) prompt, which requires the LLM to respond only with their answer without any reasoning, and (2) Zero-shot Chain-of-Though (CoT) prompt \cite{kojima2022large} (the full prompts can be found in Section~\ref{ssec:app_prompt} of the Technical Appendix). All experiments discussed are applied to both Prisoner's Dilemma and Stag Hunt.
We run all experiments on 4 LLMs (spanning both open-source and proprietary models of different sizes): \textbf{{GPT-3.5}}, \textbf{{GPT-4-Turbo}}, \textbf{{GPT-4o}}, and \textbf{{Llama-3-8B}}.


\paragraph{\textsc{Statistical Analysis.}} We test the statistical significance of all the biases. We run each experimental setup 100 times, from which we build the contingency tables required to perform the Fisher Exact Test \cite{kim2017statistical} (further details on the Fisher Exact Test can be found in Section~\ref{ssec:app_fet} of the Technical Appendix). 


\paragraph{\textsc{Alignment Analysis.}} Using the generated tables, we perform what we refer to as an alignment analysis. We compare the performance of the LLM when the experimental setup is aligned and misaligned with the LLMs' bias. Performance is assessed based on the selection of the best action given the preferred behaviours of both the acting player and fellow player (the combinations of preferred behaviours and their respective correct actions can be found in Section~\ref{ssec:pref_behav} of the Technical Appendix).

\begin{table*}[ht!]
\begin{center}
\renewcommand{\arraystretch}{1.2}
\begin{tabular}{cccccccccc}
\hline
 &
  &
 \multicolumn{4}{c}{SH} &
 \multicolumn{4}{c}{PD} \\
 &
  &
 \multicolumn{2}{c}{AO} &
 \multicolumn{2}{c}{CoT} &
 \multicolumn{2}{c}{AO} &
 \multicolumn{2}{c}{CoT} \\
 &
  &
 A $\rightarrow$ B &
 B $\rightarrow$ A &
 A $\rightarrow$ B &
 B $\rightarrow$ A &
 A $\rightarrow$ B &
 B $\rightarrow$ A &
 A $\rightarrow$ B &
 B $\rightarrow$ A \\ \hline
\multirow{2}{*}{\textbf{{GPT-3.5}}} &
 \multicolumn{1}{c|}{A} &
 \textbf{67.1} &
 0.0 &
 \textbf{53.4} &
 \multicolumn{1}{c|}{36.1} &
 \textbf{66.3} &
 0.0 &
 \textbf{56.8} &
 \textbf{50.6} \\
 &
 \multicolumn{1}{c|}{B} &
 32.9 &
 \textbf{100.0} &
 45.8 &
 \multicolumn{1}{c|}{\textbf{62.3}} &
 33.8 &
 \textbf{100.0} &
 42.4 &
 48.7 \\ \hline
\multirow{2}{*}{\textbf{{GPT-4-Turbo}}} &
 \multicolumn{1}{c|}{A} &
 25.5 &
 45.6 &
 36.1 &
 \multicolumn{1}{c|}{\textbf{53.3}} &
 25.0 &
 15.3 &
 36.8 &
 \textbf{58.8} \\
 &
 \multicolumn{1}{c|}{B} &
 \textbf{74.5} &
 \textbf{54.4} &
 \textbf{63.9} &
 \multicolumn{1}{c|}{46.8} &
 \textbf{75.0} &
 \textbf{84.8} &
 \textbf{63.3} &
 41.3 \\ \hline
\multirow{2}{*}{\textbf{{GPT-4o}}} &
 \multicolumn{1}{c|}{A} &
 25.0 &
 \textbf{73.9} &
 26.9 &
 \multicolumn{1}{c|}{\textbf{56.7}} &
 30.6 &
 \textbf{75.0} &
 25.3 &
 \textbf{67.9} \\
 &
 \multicolumn{1}{c|}{B} &
 \textbf{75.0} &
 26.1 &
 \textbf{73.1} &
 \multicolumn{1}{c|}{43.3} &
 \textbf{69.4} &
 25.0 &
 \textbf{74.8} &
 32.1 \\ \hline
\multirow{2}{*}{\textbf{{Llama-3-8b}}} &
 \multicolumn{1}{c|}{A} &
 \textbf{99.9} &
 50.0 &
 42.0 &
 \multicolumn{1}{c|}{\textbf{62.7}} &
 \textbf{100.0} &
 50.0 &
 48.7 &
 \textbf{51.6} \\
 &
 \multicolumn{1}{c|}{B} &
 0.1 &
 50.0 &
 \textbf{57.7} &
 \multicolumn{1}{c|}{36.8} &
 0.0 &
 50.0 &
 \textbf{51.2} &
 48.1 \\ \hline
\end{tabular}
\end{center}
\caption{Table showing frequency at which labels A and B are selected based on the prompted order (A first then B [A $\rightarrow$ B] or B first then A [B $\rightarrow$ A]) of the LLM. Results are shown for both games, Stag Hunt (SH) and Prisoner's Dilemma (PD). Each game is tested using Answer-Only (AO) and Zero-shot Chain-of-Thought (CoT) prompting.}
\label{tab:order_summary}
\end{table*}

\begin{figure*}[ht!]
\begin{center}
\includegraphics[width=0.45\textwidth]{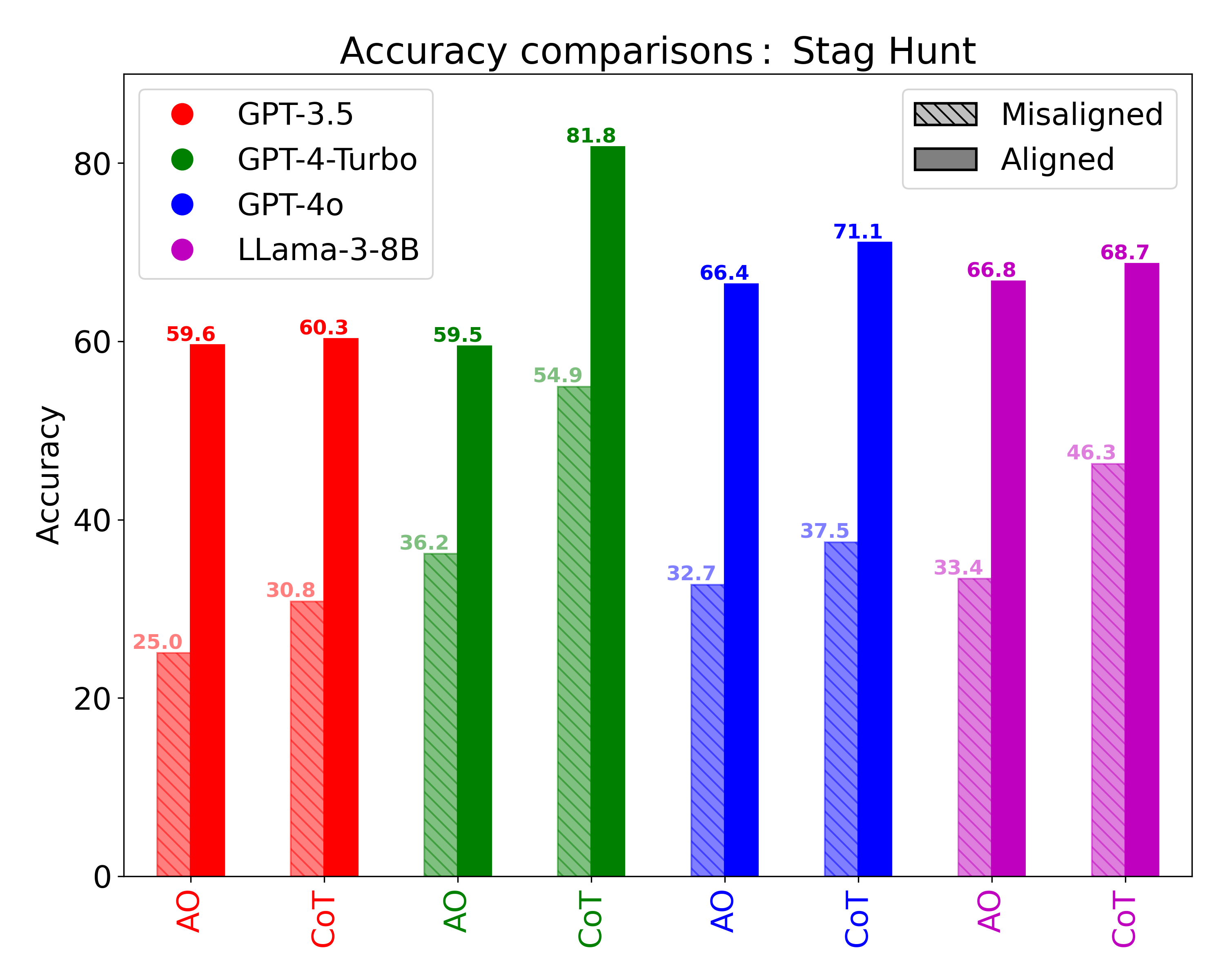}
\includegraphics[width=0.45\textwidth]{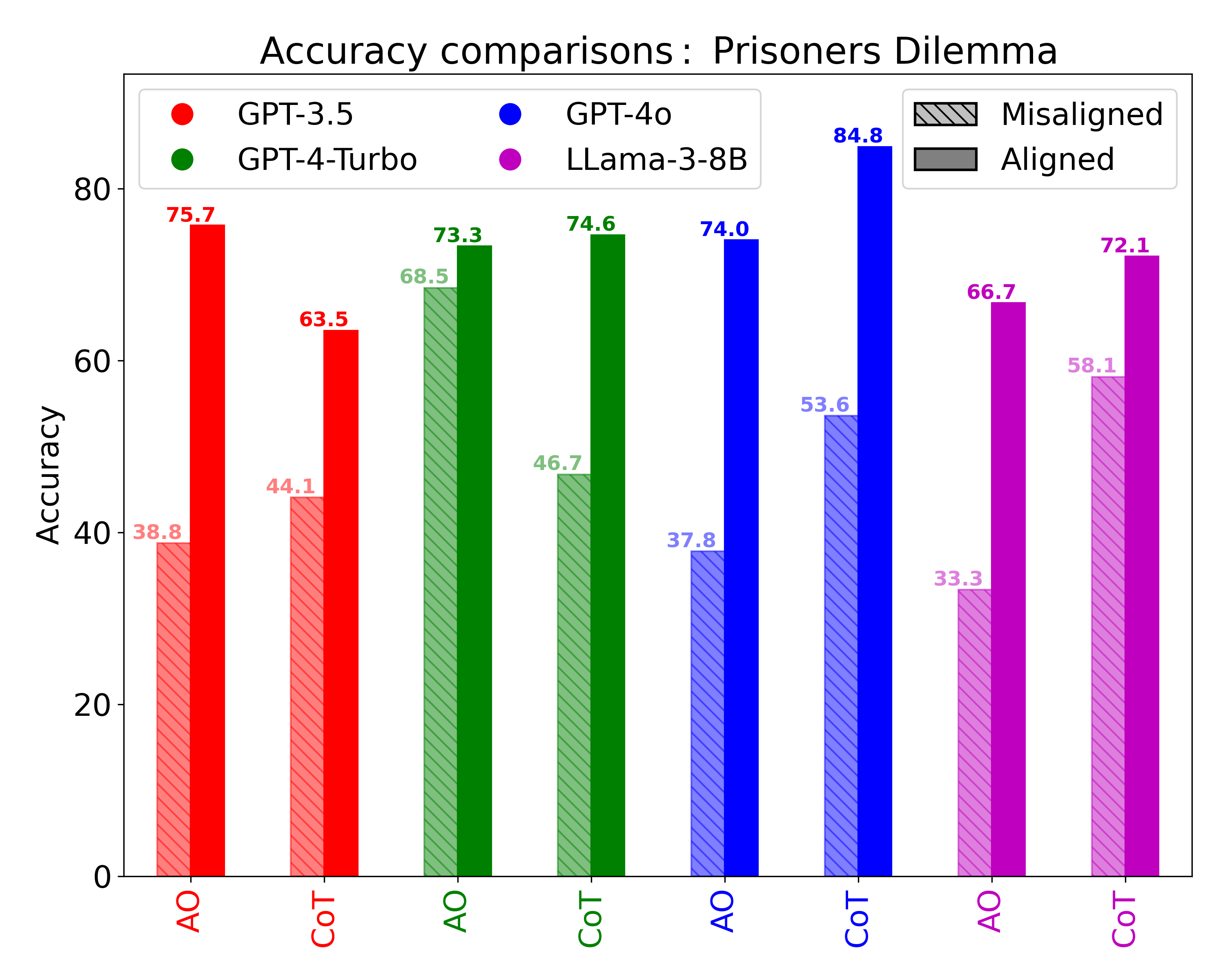}
\caption {Figure comparing the performance (measured based on the selection of the correct action given the prompted preferred behaviours) for each model under the two tested prompting methods: (1) Answer-Only (AO) and (2) Chain-of-Thought (CoT). We see that, in most experiments, CoT enables the models to achieve a higher performance in both aligned and misaligned settings. We can also consider the difference in accuracy between misalignment and alignment, namely; (LEFT) Stag Hunt - \textbf{AO:} \textbf{{34.5}}, \textbf{{23.3}}, \textbf{{33.7}}, \textbf{{33.4}} and \textbf{CoT:} \textbf{{29.5}}, \textbf{{26.9}}, \textbf{{33.6}}, \textbf{{22.4}} and (RIGHT) Prisoner's Dilemma - \textbf{AO:} \textbf{{36.9}}, \textbf{{4.8}}, \textbf{{36.2}}, \textbf{{33.3}} and \textbf{CoT:} \textbf{{19.4}}, \textbf{{27.8}}, \textbf{{31.3}}, \textbf{{14.0}}. We note that all models, except for GPT-4-Turbo, have a smaller difference in performance when using CoT prompting. A more detailed alignment analysis can be seen in Figure \ref{fig:acc} in Section~\ref{ssec:app_res_tabs} of the Technical Appendix. }
\label{fig:acc_pm}
\end{center}
\end{figure*}

\begin{table*}[ht!]
\begin{center}
\renewcommand{\arraystretch}{1.2}
\renewcommand{\arraystretch}{1.2}
\begin{tabular}{cccccccccc}
\hline
 &
  &
 \multicolumn{4}{c}{SH} &
 \multicolumn{4}{c}{PD} \\
 &
  &
 \multicolumn{2}{c}{AO} &
 \multicolumn{2}{c}{CoT} &
 \multicolumn{2}{c}{AO} &
 \multicolumn{2}{c}{CoT} \\
 &
  &
 A=C &
 B=C &
 A=C &
 B=C &
 A=C &
 B=C &
 A=C &
 B=C \\ \hline
\multirow{2}{*}{\textbf{{GPT-3.5}}} &
 \multicolumn{1}{c|}{A} &
 35.4 &
 31.8 &
 \textbf{52.4} &
 \multicolumn{1}{c|}{37.1} &
 25.0 &
 41.3 &
 46.1 &
 \textbf{61.2} \\
 &
 \multicolumn{1}{c|}{B} &
 \textbf{64.6} &
 \textbf{68.3} &
 46.2 &
 \multicolumn{1}{c|}{\textbf{62.0}} &
 \textbf{75.0} &
 \textbf{58.8} &
 \textbf{52.9} &
 38.2 \\ \hline
\multirow{2}{*}{\textbf{{GPT-4-Turbo}}} &
 \multicolumn{1}{c|}{A} &
 \textbf{58.3} &
 20.2 &
 \textbf{74.4} &
 \multicolumn{1}{c|}{14.9} &
 13.4 &
 26.9 &
 \textbf{67.0} &
 28.5 \\
 &
 \multicolumn{1}{c|}{B} &
 41.8 &
 \textbf{79.8} &
 25.6 &
 \multicolumn{1}{c|}{\textbf{85.1}} &
 \textbf{86.6} &
 \textbf{73.1} &
 33.0 &
 \textbf{71.5} \\ \hline
\multirow{2}{*}{\textbf{{GPT-4o}}} &
 \multicolumn{1}{c|}{A} &
 \textbf{75.0} &
 39.1 &
 \textbf{71.7} &
 \multicolumn{1}{c|}{12.0} &
 \textbf{75.0} &
 30.6 &
 \textbf{56.2} &
 36.9 \\
 &
 \multicolumn{1}{c|}{B} &
 25.0 &
 \textbf{60.9} &
 28.3 &
 \multicolumn{1}{c|}{\textbf{88.0}} &
 25.0 &
 \textbf{69.4} &
 43.8 &
 \textbf{63.1} \\ \hline
\multirow{2}{*}{\textbf{{Llama-3-8b}}} &
 \multicolumn{1}{c|}{A} &
 \textbf{75.0} &
 \textbf{74.9} &
 \textbf{63.3} &
 \multicolumn{1}{c|}{41.4} &
 \textbf{75.0} &
 \textbf{75.0} &
 43.5 &
 \textbf{56.9} \\
 &
 \multicolumn{1}{c|}{B} &
 25.0 &
 25.1 &
 36.4 &
 \multicolumn{1}{c|}{\textbf{58.1}} &
 25.0 &
 25.0 &
 \textbf{56.3} &
 43.0 \\ \hline
\end{tabular}
\end{center}
\caption{Table showing the frequency at which labels A and B are selected based on the prompted payoffs associated with the labels (A is Cooperative [A=C] or B is Cooperative [B=C]) of the LLM. Results are shown for both games, Stag Hunt (SH) and Prisoner's Dilemma (PD). Each game is tested using Answer-Only (AO) and Zero-shot Chain-of-Thought (CoT) prompting.}
\label{tab:payoff_summary}
\end{table*}

\begin{table}[ht!]
\begin{center}
\renewcommand{\arraystretch}{1.2}
\begin{tabular}{ccccccc}
\hline
            &           &            & \multicolumn{2}{c}{SH}               & \multicolumn{2}{c}{PD}    \\
            &           &            & AO      & CoT                & AO      & CoT      \\ \hline
\multirow{4}{*}{\rotatebox[origin=c]{45}{\textbf{{GPT-3.5}}}} &
 \multirow{2}{*}{SG} &
 \multicolumn{1}{c|}{A} &
 17.1 &
 \multicolumn{1}{c|}{17.0} &
 16.3 &
 29.6 \\
            &           & \multicolumn{1}{c|}{B} & \textbf{82.9} & \multicolumn{1}{c|}{\textbf{82.6}} & \textbf{83.8} & \textbf{69.9} \\ \cline{4-7} 
            & \multirow{2}{*}{CG} & \multicolumn{1}{c|}{A} & 50.0 & \multicolumn{1}{c|}{\textbf{72.5}} & 50.0 & \textbf{77.8} \\
            &           & \multicolumn{1}{c|}{B} & 50.0     & \multicolumn{1}{c|}{25.5}     & 50.0     & 21.2     \\ \hline
\multirow{4}{*}{\rotatebox[origin=c]{45}{\textbf{{GPT-4-Turbo}}}} &
 \multirow{2}{*}{SG} &
 \multicolumn{1}{c|}{A} &
 33.4 &
 \multicolumn{1}{c|}{42.7} &
 26.9 &
 46.5 \\
            &           & \multicolumn{1}{c|}{B} & \textbf{66.6} & \multicolumn{1}{c|}{\textbf{57.3}} & \textbf{73.1} & \textbf{53.5} \\ \cline{4-7} 
            & \multirow{2}{*}{CG} & \multicolumn{1}{c|}{A} & 37.8     & \multicolumn{1}{c|}{46.7}     & 13.4     & 49.0     \\
            &           & \multicolumn{1}{c|}{B} & \textbf{62.3} & \multicolumn{1}{c|}{\textbf{53.3}} & \textbf{86.6} & \textbf{51.0} \\ \hline
\multirow{4}{*}{\rotatebox[origin=c]{45}{\textbf{{GPT-4o}}}} & \multirow{2}{*}{SG} & \multicolumn{1}{c|}{A} & 48.9     & \multicolumn{1}{c|}{36.2}     & 50.0     & 44.0     \\
            &           & \multicolumn{1}{c|}{B} & \textbf{51.1} & \multicolumn{1}{c|}{\textbf{63.8}} & 50.0     & \textbf{56.0} \\ \cline{4-7} 
            & \multirow{2}{*}{CG} & \multicolumn{1}{c|}{A} & 50.0     & \multicolumn{1}{c|}{47.5}     & \textbf{55.6} & 49.1     \\
            &           & \multicolumn{1}{c|}{B} & 50.0 & \multicolumn{1}{c|}{\textbf{52.5}} & 44.4     & \textbf{50.9} \\ \hline
\multicolumn{1}{l}{\multirow{4}{*}{\rotatebox[origin=c]{45}{\textbf{{Llama-3-8b}}}}} &
 SG &
 \multicolumn{1}{c|}{A} &
 \textbf{100.0} &
 \multicolumn{1}{c|}{\textbf{64.8}} &
 \textbf{100.0} &
 \textbf{68.5} \\
\multicolumn{1}{l}{}  &           & \multicolumn{1}{c|}{B} & 0.0      & \multicolumn{1}{c|}{34.8}     & 0.0      & 31.3     \\ \cline{4-7} 
\multicolumn{1}{l}{}  & \multirow{2}{*}{CG} & \multicolumn{1}{c|}{A} & 49.9     & \multicolumn{1}{c|}{39.9}     & 50.0     & 31.9     \\
\multicolumn{1}{l}{}  &           & \multicolumn{1}{c|}{B} & \textbf{50.1} & \multicolumn{1}{c|}{\textbf{59.7}} & 50.0     & \textbf{68.0} \\ \hline
\end{tabular}
\end{center}
\caption{Table showing frequency at which labels A and B are selected based on the prompted behaviours, Self-Gain (SG) and Common-Gain (CG), of the LLM. Results are shown for both games, Stag Hunt (SH) and Prisoner's Dilemma (PD). Each game is tested using Answer-Only (AO) and Zero-shot Chain-of-Thought (CoT) prompting.}
\label{tab:behaviour_summary}
\end{table}

\section{Main Results}
\label{sec:results}

We tested a total of 4 LLMs under a total of 16 different prompt configurations (where we varied the position, payoff, and preferred behaviours in the prompt). Furthermore, we run each prompt configuration under 2 different prompting schemes, AO and CoT. The frequency tables for each bias are shown in Tables \ref{tab:order_summary}, \ref{tab:payoff_summary}, and \ref{tab:behaviour_summary}. The results of the statistical analysis can be found in Figure \ref{fig:stat_sig} and the alignment analyst results in Figure \ref{fig:acc_pm}. In the following paragraphs, we discuss these results in depth. 

\paragraph{\textsc{Positional Bias.}} In Figure \ref{fig:stat_sig}, we can see that the positional bias is particularly \textbf{strong in GPT-3.5} when using the AO prompt. It becomes significantly weaker when asked to reason over the task first when using the CoT prompt. \textbf{GPT-4-Turbo}, on the other hand, shows an {\textbf{overall much weaker} bias towards the position of the action labels, for both prompting methods. Interestingly, the newer {\textbf{GPT-4o regresses} and shows a {\textbf{strong} positional bias under both prompting methods. Lastly, \textbf{Llama-3-8B}, much like GPT-3.5, shows a \textbf{strong positional bias} under AO prompting and a much weaker bias under CoT prompting. More specifically, in Table \ref{tab:order_summary}, we see that GPT-3.5 tends to select the action in the first position more frequently. Conversely, GPT-4-Turbo and GPT-4o tend to select the action in the second position more frequently. Llama-3-8B, under AO prompting, selects the first position more frequently and the second position under CoT prompting.

\paragraph{\textsc{Payoff Bias.}} In Figure \ref{fig:stat_sig}, we can see that both \textbf{GPT-3.5 and Llama-3-8B} show either a \textbf{very weak or insignificant} payoff bias for both prompting methods. Interestingly, both models tend to (1) maximise the common-gain in Stag Hunt and (2) maximise the self-gain in Prisoner's Dilemma. This can be seen in Table \ref{tab:payoff_summary}. Looking at Figure \ref{fig:stat_sig} again, we see that the payoff bias is \textbf{strong in both GPT-4-Turbo and GPT-4o}. It is the strongest bias in GPT-4-Turbo for both prompting methods in both games, whereas for GPT-4o the payoff bias is weaker under CoT prompting in Prisoner's Dilemma. Presumably, the reason for this is that the maximum possible self-gain and common-gain in Prisoner's Dilemma is less than in Stag Hunt and is, therefore, a weaker signal. In table \ref{tab:payoff_summary} (and Table~\ref{tab:cons_self_summary} in Section~\ref{ssec:app_res_tabs} of the Technical Appendix) we note that both GPT-4-Turbo and GPT-4o tend to maximise the common-gain more frequently. While in previous works, this behaviour has led to conclusions such as ``LLMs have a propensity to be cooperative'' \cite{xu2023magic, brookins2023playing}, our results suggest that this phenomenon is not indicative of ``cooperative behaviour'' and instead is a result of a skewed attention towards the action that leads to maximum possible gains, despite it not always being the optimal choice. 

\paragraph{\textsc{Behavioural Bias.}} In Figure \ref{fig:stat_sig}, we can see that both \textbf{GPT-4-Turbo and GPT-4o} are \textbf{weakly affected} by the behavioural bias, whereas, both \textbf{GPT-3.5 and Llama-3-8B} are \textbf{strongly affected}. Specifically, GPT-3.5 is weakly biased when using the AO prompt and strongly biased when using the CoT prompt. Llama-3-8B shows a different pattern where it is strongly biased when using the AO prompt and weakly biased when using the CoT prompt. However, in Prisoner's Dilemma, it is still the strongest bias under the CoT prompt. In Table \ref{tab:behaviour_summary}, we observed that GPT-3.5 tends to select action label A when prompted to prioritise Common-Gain and action label B when prompted to prioritise Self-Gain. Again, Llama-3-8B shows a different pattern, where it tends to select action label A when prompted to prioritise Self-Gain and action label B when prompted to prioritise Common-Gain.

\paragraph{\textsc{High-level Analysis.}} Now that the fine-grained details of each identified bias have been discussed, let us consider the high-level effects these biases have on the performance of LLMs playing these games. In Figure \ref{fig:acc_pm}, it can be seen that in all models, under both prompting methods, when the bias of the LLM and the experimental setup are misaligned, the performance is much worse. Specifically, \textbf{{GPT-3.5}}, \textbf{{GPT-4-Turbo}}, \textbf{{GPT-4o}}, and \textbf{{Llama-3-8B}} show an average performance drop, over both prompting methods, of \textbf{{32\%}}, \textbf{{25\%}}, \textbf{{34\%}}, and \textbf{{28\%}}, respectively in Stag Hunt, and \textbf{{28\%}}, \textbf{{16\%}}, \textbf{{34\%}}, and \textbf{{24\%}} respectively in Prisoner's Dilemma. It's worth noting that employing \textbf{CoT prompting} \textbf{lessens} the performance drop due to \textbf{bias} misalignment by \textbf{31.1\%}, \textbf{6.93\%}, and \textbf{45.39\%} in \textbf{GPT-3.5}, \textbf{GPT-4o}, and \textbf{Llama-3-8B} respectively, and increases the effect of the bias in \textbf{GPT-4-Turbo} by \textbf{247.22\%}. This is because there is a significant relationship between the performance drop and the strength of the biases experienced by the models (see Figure \ref{fig:sig_vs_diff} for the specifics) and it can be seen in Figure \ref{fig:stat_sig} that GPT-4-Turbo's biases strengthen with CoT prompting. All models, except GPT-4-Turbo under CoT prompting, perform equally or better in the Prisoner's Dilemma compared to the Stag Hunt. The suspected that in Prisoner's Dilemma the action which maximises self-gain is the correct action 75\% of the time (in Stag Hunt it is 50\%, refer to Section~\ref{ssec:pref_behav} of the Technical Appendix for more detail), for all experimental setups, and that GPT-4-Turbo is the only model which is primarily biased to select the action which maximises the common-gain (see Table \ref{tab:payoff_summary} and Table~\ref{tab:cons_self_summary} in the Technical Appendix) which causes it to select the incorrect action more often.

\section{Additional Experiments and Results}

\begin{figure}[ht!]
\includegraphics[width=0.45\textwidth]{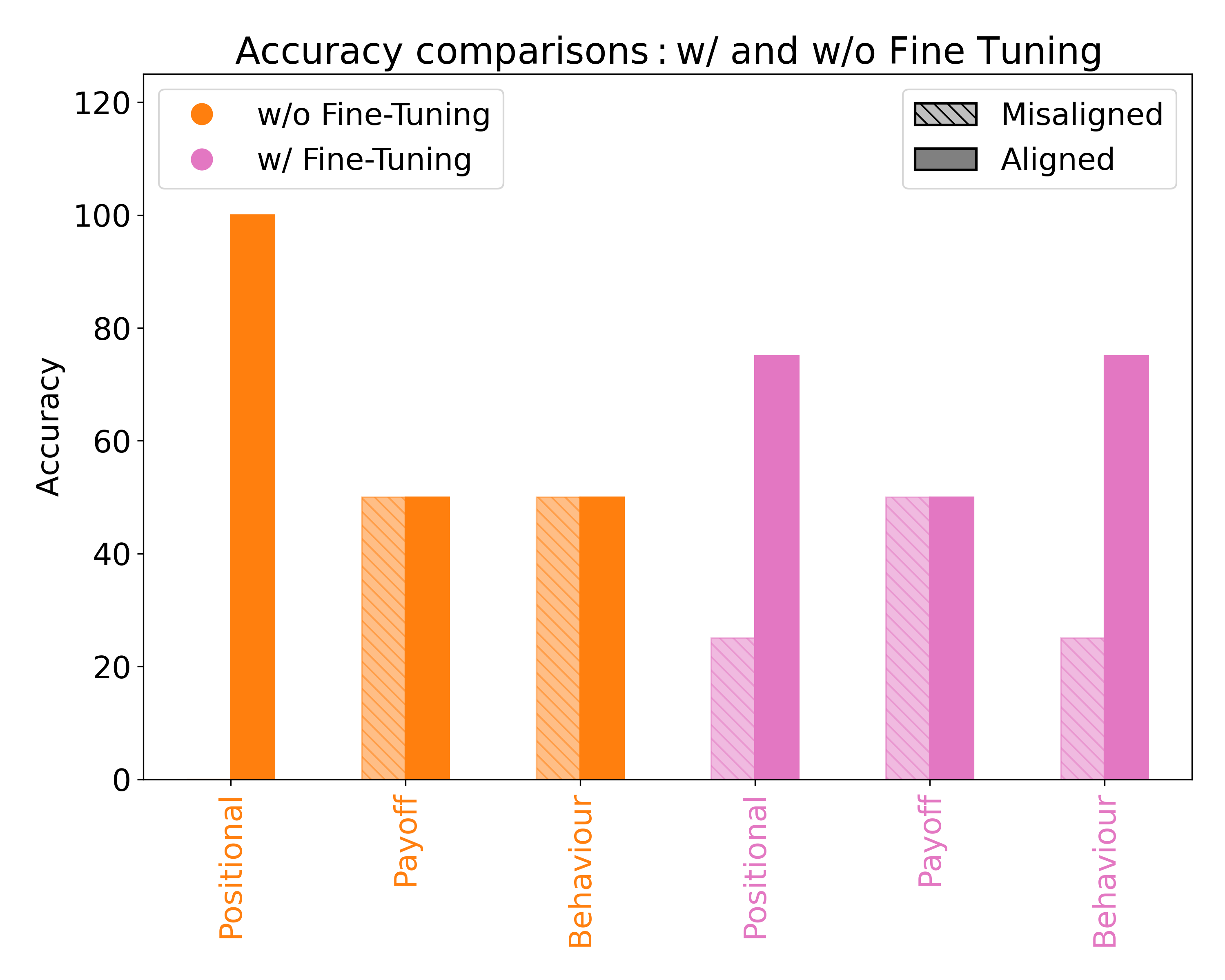}
\caption{Figure comparing the performance (aligned and misaligned for each systematic bias) of LLama-3-8b (w/o Fine-Tuning) and LLama-3-8b-Instruct (w/ Fine-Tuning) using the Answer-Only (AO) prompting scheme.}
\label{fig:add_res_ft_acc}
\end{figure}

\begin{figure*}[ht!]
\begin{center}
\includegraphics[width=0.45\textwidth]{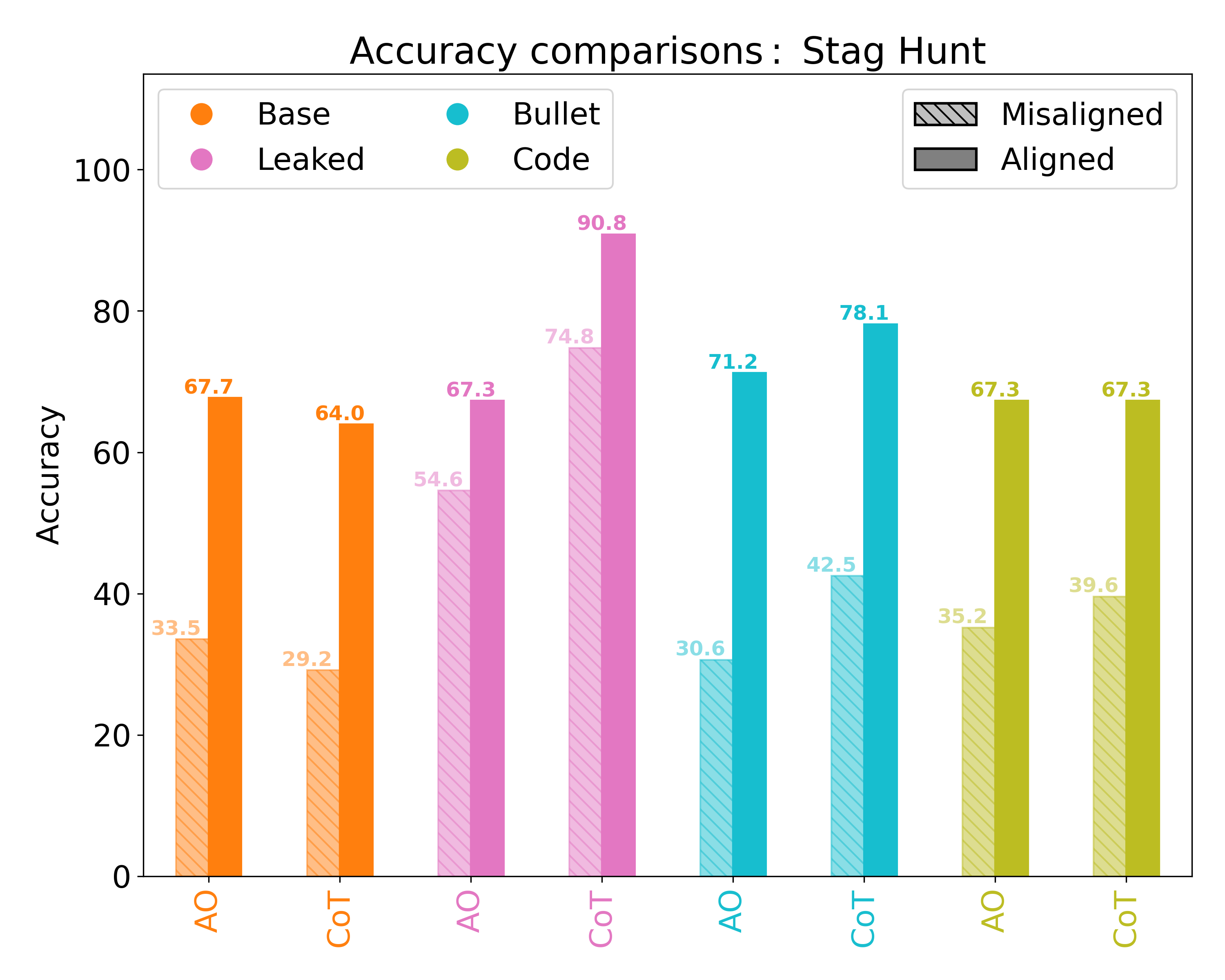}
\includegraphics[width=0.45\textwidth]{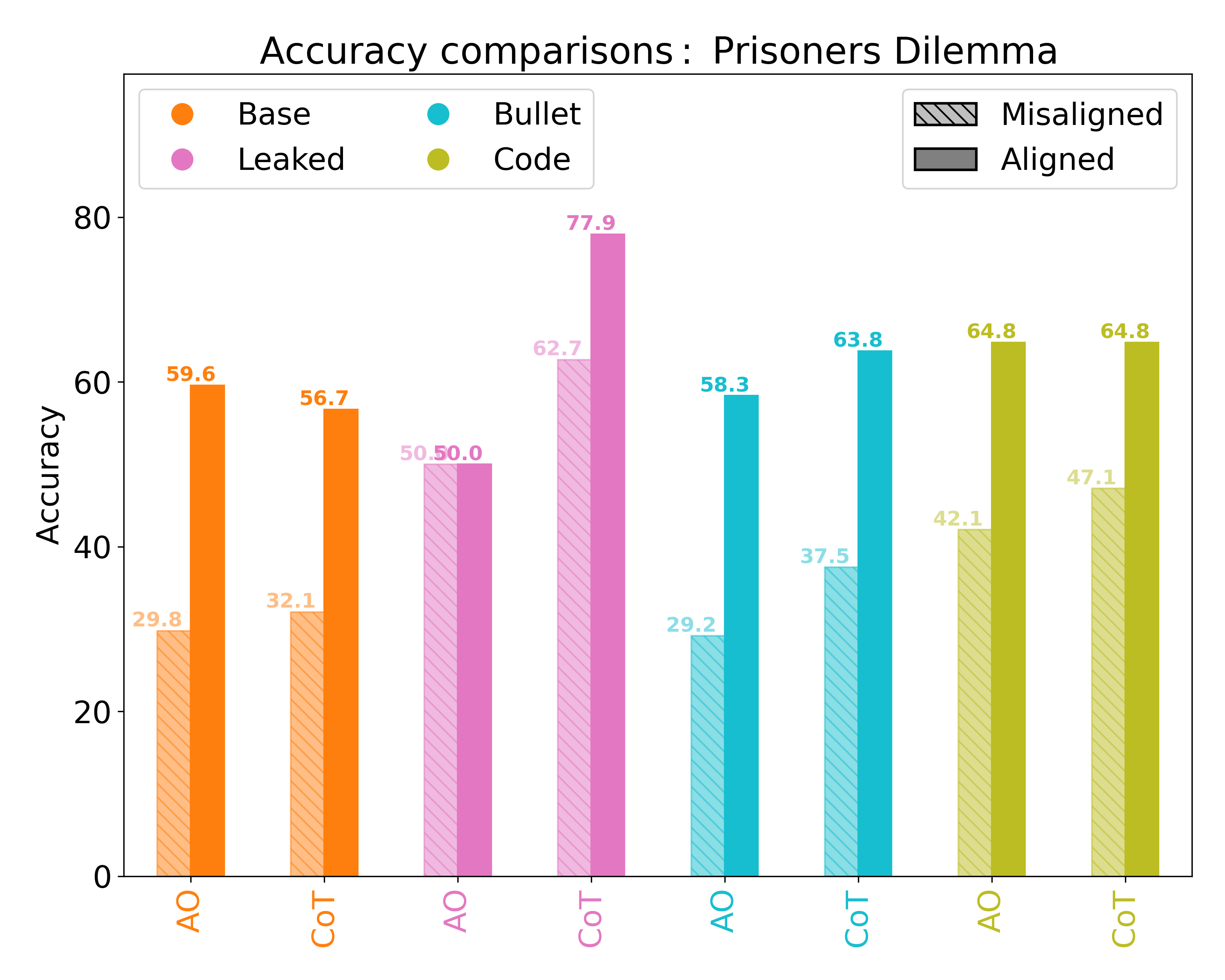}
\caption{Figure comparing the performance (misaligned vs aligned) of GPT-4o under different prompts. This is done for both tested prompting schemes: (1) Answer-Only (AO) and (2) Chain-of-Thought (CoT).}
\label{fig:add_res_acc}
\end{center}
\end{figure*}

To further understand the behaviour observed in the LLMs, we perform a set of additional experiments. The findings are briefly discussed below, while more details can be found in Section~\ref{sec:add_results} of the Technical Appendix.

\subsection{Effect of Fine-Tuning}

We ran a set of experiments using LLama-3-8b (w/o Fine-Tuning) and LLama-3-8b-Instruct (w/ Fine-Tuning) to investigate the effect instruction-fine-tuning has on the identified biases. In Figure~\ref{fig:add_res_ft_acc} we see that LLama-3-8b was primarily guided by the order in which the action labels were presented in the prompt, whereas LLama-3-8b-Instruct is equally guided by both the order and prompted preferred behaviours (this result can be observed in Figure~\ref{fig:add_res_ft_stat} as well). However, the overall effect of the biases is weakened when the LLM is fine-tuned. Our results show that fine-tuning LLMs reduces the average significance of the biases, however, they also show that fine-tuning increases the behavioural bias in LLama-3-8b (where it was null to begin with), indicating fine-tuning may not be a sufficient remedy for this issue.

\subsection{Effect of Prior Knowledge}

To investigate the effect an LLM's prior knowledge has on the identified biases, we alter the base prompt used in the main set of experiments to include the names of the games (referred to as the leaked prompt) i.e. Prisoner's Dilemma and Stag Hunt. In Figure~\ref{fig:add_res_acc} we see that when prompting GPT-4o with the names of the games in the prompt, the effect of the biases weakens (since the performance drop when misaligned is less). This result can also be observed in Figure~\ref{fig:add_res_sig} as well. This drop is likely due to GPT-4o relying more on its prior knowledge about these games, rather than reasoning from scratch. Since GPT-4o is likely to have encountered these games and their solutions during training, it can solve the different instances with a higher degree of accuracy. 
 
\subsection{Effect of Prompt Variations}

Here we aim to check whether these biases, which present themselves in the main set of experiments, are specific to the base prompt used or inherent in the LLM itself. To do this, we evaluate GPT-4o under different prompts, which we refer to as (1) bullet prompt and (2) code prompt (both found in Section~\ref{sec:add_results} of the Technical Appendix). In Figure~\ref{fig:add_res_acc} we see that the biases present themselves under all these prompts (GPT-4o experiences a performance drop when misaligned for all prompts), albeit not equally. We also see in Figure~\ref{fig:add_res_sig} that the biases are still significant under all prompts. This shows that by changing the prompt, one can alter how the biases affect the LLM's decision-making abilities, it also shows that one cannot simply expect to rid LLMs of their biases by changing the prompt alone, indicating such biases are inherent.

\subsection{Humans vs AI}

We compare the frequency at which humans and an LLM (GPT-4o) select specific actions in a Stag Hunt-like game. We found that humans (data provided by \citet{belloc2019intuition}) select action label B (cooperative action) more frequently than action label A (selfish action). We also see that GPT-4o, under the same setting as the humans, selects action label B more frequently. We then test GPT-4o under a different setting (where B is selfish and A is cooperative) and note that it still selects action label B more frequently (see Figure~\ref{fig:prelim} in Section~\ref{sec:add_results} of the Technical Appendix for more details). This suggests that GPT-4o is greatly influenced by its biases, rather than using logical reasoning to solve the problem and that we should be careful when comparing the performance of LLMs and humans in such tasks.

\section{Conclusion}

We performed an in-depth analysis of SOTA LLMs, showing that they are significantly affected by at least one of the 3 identified biases (see Sections \ref{sec:meth} and \ref{sec:results} for details) (1) positional, (2) payoff, and (3) behavioural, irrespective of the prompt used. These biases cause notable changes in the LLMs' performances under different configurations, even though the underlying task remains unchanged. We show that models GPT-3.5, GPT-4-Turbo, GPT-4o, and Llama-3-8B exhibit an average performance drop of 31\%, 21\%, 34\%, and 27\%, respectively across the tested games. We highlight the importance of taking special care of how these biases affect performance when evaluating LLMs in game-theoretic tasks. 

Further, we observe that, while GPT-4o outperforms all of the other tested models on the standard LLM evaluation metrics \cite{OpenAI}, it suffers the worst in terms of performance drop suggesting that these metrics of success are not sufficient indicators of how well LLMs will perform in game theoretic tasks, especially when considering the effects of these systematic biases. Using CoT prompting generally reduces the impact of biases in most models and helps mitigate performance drops, suggesting that further work in prompt engineering might address these biases to an extent. 

However, this effect is not consistent across all models; for instance, GPT-4-Turbo exhibits stronger biases with CoT prompting, leading to greater performance drops. This indicates that the solutions effective for some models may not be a general one across all models. We believe that the root issue of these systematic biases is the inherently limited reasoning capabilities of LLMs. Further work into improving these abilities will resolve the current limitations of LLMs on their systematic biases to make these decisions.

\bibliography{custom}


\newpage
\onecolumn
\section*{Reproducibility Checklist}

This paper:
\begin{itemize}
  \item Includes a conceptual outline and/or pseudocode description of AI methods introduced (NA)
  \item Clearly delineates statements that are opinions, hypothesis, and speculation from objective facts and results (yes)
  \item Provides well marked pedagogical references for less-familiare readers to gain background necessary to replicate the paper (yes)
\end{itemize}

\noindent Does this paper make theoretical contributions? (no) \\
\noindent If yes, please complete the list below.

\begin{itemize}
  \item All assumptions and restrictions are stated clearly and formally. (yes/partial/no)
  \item All novel claims are stated formally (e.g., in theorem statements). (yes/partial/no)
  \item Proofs of all novel claims are included. (yes/partial/no)
  \item Proof sketches or intuitions are given for complex and/or novel results. (yes/partial/no)
  \item Appropriate citations to theoretical tools used are given. (yes/partial/no)
  \item All theoretical claims are demonstrated empirically to hold. (yes/partial/no/NA)
  \item All experimental code used to eliminate or disprove claims is included. (yes/no/NA)
\end{itemize}

\noindent Does this paper rely on one or more datasets? (no) \\
\noindent If yes, please complete the list below.

\begin{itemize}
  \item A motivation is given for why the experiments are conducted on the selected datasets (yes/partial/no/NA)
  \item All novel datasets introduced in this paper are included in a data appendix. (yes/partial/no/NA)
  \item All novel datasets introduced in this paper will be made publicly available upon publication of the paper with a license that allows free usage for research purposes. (yes/partial/no/NA)
  \item All datasets drawn from the existing literature (potentially including authors’ own previously published work) are accompanied by appropriate citations. (yes/no/NA)
  \item All datasets drawn from the existing literature (potentially including authors’ own previously published work) are publicly available. (yes/partial/no/NA)
  \item All datasets that are not publicly available are described in detail, with explanation why publicly available alternatives are not scientifically satisficing. (yes/partial/no/NA)
\end{itemize}

\noindent Does this paper include computational experiments? (yes)\\
\noindent If yes, please complete the list below.

\begin{itemize}
  \item Any code required for pre-processing data is included in the appendix. (no).
  \item All source code required for conducting and analyzing the experiments is included in a code appendix. (no)
  \item All source code required for conducting and analyzing the experiments will be made publicly available upon publication of the paper with a license that allows free usage for research purposes. (yes)
  \item All source code implementing new methods have comments detailing the implementation, with references to the paper where each step comes from (yes)
  \item If an algorithm depends on randomness, then the method used for setting seeds is described in a way sufficient to allow replication of results. (NA)
  \item This paper specifies the computing infrastructure used for running experiments (hardware and software), including GPU/CPU models; amount of memory; operating system; names and versions of relevant software libraries and frameworks. (no)
  \item This paper formally describes evaluation metrics used and explains the motivation for choosing these metrics. (yes)
  \item This paper states the number of algorithm runs used to compute each reported result. (yes)
  \item Analysis of experiments goes beyond single-dimensional summaries of performance (e.g., average; median) to include measures of variation, confidence, or other distributional information. (yes)
  \item The significance of any improvement or decrease in performance is judged using appropriate statistical tests (e.g., Wilcoxon signed-rank). (yes)
  \item This paper lists all final (hyper-)parameters used for each model/algorithm in the paper’s experiments. (NA)
  \item This paper states the number and range of values tried per (hyper-) parameter during development of the paper, along with the criterion used for selecting the final parameter setting. (NA)
\end{itemize}

\clearpage
\appendix
\section{Background Information}
\label{sec:further}

To provide more comprehensive information for broader readers, the following section describes Game Theory and the Fisher Exact Test in further detail. 

\subsection{Game Theory}
\label{ssec:app_game}

Game Theory is the study of how the choices of interacting agents with specific preferences produce outcomes, intentional and not \cite{sep-game-theory}. Game theory is currently applied to many existing real-world tasks in domains such as economics, politics, and psychology \cite{martin2017course}. Game theory models assume that the interacting agents make rational choices, which can be modelled as follows \cite{osborne1994course}:

\begin{enumerate}
  \item A set of actions $A$ from which the agents select their choice.
  \item A set of possible consequences $C$ to action set $A$.
  \item A function $g: A \rightarrow C$ that maps actions to consequences.
  \item A preference relation\footnote{Consider sources \cite{osborne1994course} and \cite{albouy2004preference} for details on the concept of preference relations.} $\succeq$ on set $C$. Note that $\succeq$ can be defined by a utility/payoff function $U: C \rightarrow \mathbb{R}$ where $x \succeq y \iff U(x) \geq U(y)$.
\end{enumerate}

\noindent Therefore, a rational agent chooses $a^* \in A$ if $g(a^*) \succeq g(a)$ for all $a \in A$. 

These situations, in which rational agents interact with each other by taking action simultaneously, are referred to as strategic games \cite{osborne1994course}. The following defines a strategic game:

\begin{enumerate}
  \item A finite set of players $N$.
  \item A nonempty set of actions available to $\textrm{agent}_i$ $A_i$
  \item A preference relation for each $\textrm{agent}_i$ $\succeq_i$ on set $A = A_j \times A_j$ for $j \in N$. This is what distinguishes a strategic game from a decision problem.
\end{enumerate}

While many different qualifications exist for these games, this article focuses on non-zero-sum games. The reason for this is as follows; In zero-sum games, an optimal solution can always be found due to its strictly competitive nature, which is not a fair representation of rational agent interactions for many important real-world scenarios. This is not the case for non-zero-sum games, which can have both competitive and cooperative elements. Popular examples of such games are Stag Hunt and Prisoner's Dilemma, which will be the focus of this research. The contingency tables of which can be seen in Table \ref{tab:contingency_app}.

\begin{table}[ht]
\centering
\renewcommand{\arraystretch}{1.2}
\begin{tabular}{cccccc}
 &
 Stag &
 Hare &
  &
 Quiet &
 Confess \\ \cline{2-3} \cline{5-6} 
\multicolumn{1}{c|}{Stag} &
 \textbf{{5, 5}} &
 \multicolumn{1}{c|}{0, 4} &
 \multicolumn{1}{c|}{Quiet} &
 2, 2 &
 \multicolumn{1}{c|}{0, 3} \\
\multicolumn{1}{c|}{Hare} &
 4, 0 &
 \multicolumn{1}{c|}{\textbf{{2, 2}}} &
 \multicolumn{1}{c|}{Confess} &
 3, 0 &
 \multicolumn{1}{c|}{\textbf{{1, 1}}} \\ \cline{2-3} \cline{5-6} 
\end{tabular}
\caption{Payoff matrices for (LEFT) Stag Hunt and (RIGHT) Prisoner's Dilemma. The Nash Equilibrium(s) for each is shown in \textbf{{red}} for each game.}
\label{tab:contingency_app}
\end{table}

A key concept which is used to find an equilibrium in non-zero-sum games is the Nash Equilibrium \cite{chatterjee2004two}. A Nash Equilibrium is a solution to the game if no player can improve their outcome by unilaterally changing their decision. More formally \cite{osborne1994course}:

A Nash Equilibrium of a strategic game $(N, A, \succeq)$ is action $a^* \in A$ where $(a^*_i, a^*_{-i}) \succeq (a_{i}, a^*_{-i})$ for all $a_i \in A_i$ and $i \in N$.

\paragraph{Stag Hunt:} The Stag Hunt, a prototype of the social contract \cite{skyrms2001stag}, is a story of two hunters who can each hunt a hare on their own but have to work together to hunt a stag. The hunters have to decide on what action to take, without communicating with one another, based on whether they believe that their fellow hunter will choose to cooperate and hunt the stag or if they will choose to defect and hunt the hare. In formal game theory terms, it is a strategic game $(N, A, \succeq)$, where $N=2$, $A \in (\textrm{Stag}, \textrm{Hare})$, and $\succeq$ is defined by the payoff function represented as a payoff matrix in Table \ref{tab:contingency_app}. 

This game has two Nash Equilibrium (seen in \textbf{bold} in Table \ref{tab:contingency_app}); 

\begin{itemize}
  \item If the hunter believes that their fellow hunter will hunt the stag, there is no better option for them than to hunt the stag as well. Specifically, $a^* = \textrm{Stag}$ since $(\textrm{Stag}, \textrm{Stag}) \succeq (\textrm{Hare}, \textrm{Stag})$.
  \item If the hunter believes that their fellow player will hunt the hare, there is no better option for them than to hunt the hare as well. Specifically, $a^* = \textrm{Hare}$ since $(\textrm{Hare}, \textrm{Hare}) \succeq (\textrm{Stag}, \textrm{Hare})$.
\end{itemize}

\paragraph{Prisoner's Dilemma:} The Prisoner's Dilemma, an illustration of a conflict between selfish and cooperative behaviour, is a story which tells of two prisoners both faced with the same choices; (1) To confess or (2) to remain silent. If both confess, they both spend 2 years in prison, if they both remain silent, they bother to spend 1 year in prison, and if one confesses and the other remains silent, the first spends no years in prison while the other spends 3. In formal game theory terms, it is a strategic game $(N, A, \succeq)$, where $N=2$, $A \in (\textrm{Quiet}, \textrm{Confess})$, and $\succeq$ is defined by the payoff function represented as a payoff matrix in Table \ref{tab:contingency_app}. 

Unlike the Stag Hunt, there is only one Nash Equilibrium (seen in \textbf{bold} in Table \ref{tab:contingency_app}) which is to always Confess. Specifically, $a^* = \textrm{Confess}$ since $(\textrm{Confess}, \textrm{Quiet}) \succeq (\textrm{Quiet}, \textrm{Quiet})$ and $(\textrm{Confess}, \textrm{Confess}) \succeq (\textrm{Quiet}, \textrm{Confess})$.

\subsection{Fisher Exact Test}
\label{ssec:app_fet}

The Fisher Exact Test is used to analyse the statistical significance of the relationship between the rows and the columns of contingency tables \cite{kim2017statistical}. Specifically, the null hypothesis is that the columns and rows are independent \cite{mcdonald2009handbook}. Following this, the Fisher Exact test is used to calculate the p-value and for $p < 0.05$ (the null hypothesis has less than a 5\% chance of being true) we reject the null hypothesis. Typically, the Fisher Exact Test is used for smaller sample sizes but is valid for all sizes \cite{kim2017statistical}. It is one of the Exact Tests since the calculation of the p-value does not rely on approximations \cite{mcdonald2009handbook}. 

The Fisher Exact Test uses the hypergeometric distribution to calculate the p-values, which takes on the general form \cite{hoffman2015biostatistics};

\begin{equation*}
\begin{split}
  &P(A=r, A'=(n-r)) = \frac{\binom{X}{r} \binom{N-X}{n-r}}{\binom{N}{n}} \\
  &= \frac{\left(\frac{X!}{r!(X-r)!}\right)\left(\frac{(N-X)!}{(n-r)!(N-X-n+r)!}\right)}{\frac{N!}{n!(N-n!)}} 
\end{split}
\end{equation*}

where $N$ is the total population of objects, $X \in N$ have a specific characteristic $A$ and $(N-X) \in X$ do not, and $n$ is the sample size drawn from $N$. Essentially, the above is calculating the probability that $r$ samples have a specific characteristic from the $n$ samples drawn.

Now, given the example contingency Table \ref{tab:example_cont_table}, where $N$ in the sample population, where $c_1 \in N$ has characteristic $A$ and $c_2 \in N$ has characteristic $A'$. Then, from $N$, $r_1 \in N$ samples are drawn and $a \in r$ samples have characteristic $A$. Then to calculate the 2-sided p-value, the Hypergeometric Distribution function is used to calculate the probabilities that $x \in r_1$ where $0<x<a$ have characteristic $A$. These probabilities are finally summed together. Specifically;

\begin{equation*}
  FET_{2-sided} = \sum^{a}_{x=0}P(A=x, A'=(r_1-x))
\end{equation*}

\begin{table}[ht]
\caption{Example of 2x2 contingency Table.}
\label{tab:example_cont_table}

\begin{center}
\begin{tabular}{cccc}
       & A             & A'            & Total \\ \hline
In sample   & \multicolumn{1}{c|}{a}  & \multicolumn{1}{c|}{b}  & $r_1$ \\
Not in sample & \multicolumn{1}{c|}{c}  & \multicolumn{1}{c|}{d}  & $r_2$ \\ \hline
       & \multicolumn{1}{c|}{$c_1$} & \multicolumn{1}{c|}{$c_2$} & N   \\ \hline
\end{tabular}
\end{center}
\end{table}

\section{Experiment Details}
\label{sec:app_exp_details}

For the reader's interest, the following section provides additional details on the prompted preferred behaviour of the LLMs as well as the base prompts used during the main set of experiments.

\subsection{Experiment Design}

The observed output of the LLMs is affected by the models' temperature (T) (A parameter that controls the randomness of the generated text by increasing the entropy of the output token distribution). Since we are interested in the models' underlying behaviour, independent of this randomness, we perform all analyses on the models' top token (the token with the highest probability associated with it). Therefore, for AO prompting, we test at T=1.0, a commonly used default value for temperature. However, since zero-shot CoT prompts the LLM to reason, it would be amiss to not take into consideration the effect of the random sampling while generating the reasoning on which the LLM conditions their final choice. Therefore, for zero-shot CoT prompting, we analyse the results over low (T=0.0), medium (T=0.5), and high (T=1.0) temperatures. 

\subsection{Behaviour Preferences}
\label{ssec:pref_behav}

We test the LLMs under different combinations of preferred behaviours, namely; the {acting player (AP)} will be prompted with both the preferred behaviours (prioritise \textbf{{Common-Gain}} or prioritise \textbf{{Self-Gain}}) of their {fellow player (FP)} as well as their own ({AP}, {FP} $= \{CG, SG\}$). The following states which action the acting agent should select based on the preferred behaviours of both players:

\begin{tcolorbox}[breakable,colframe=black!75!black,title=Stag Hunt: Preferred Behaviours]
Acting Agent: \textbf{{Common-Gain}} + Fellow Agent: \textbf{{Common-Gain}} $\Rightarrow$ Action $=$ \textbf{{Common-Gain}}

Acting Agent: \textbf{{Common-Gain}} + Fellow Agent: \textbf{{Self-Gain}} $\Rightarrow$ Action $=$ \textbf{{Self-Gain}}

Acting Agent: \textbf{{Self-Gain}} + Fellow Agent: \textbf{{Common-Gain}} $\Rightarrow$ Action $=$ \textbf{{Common-Gain}}

Acting Agent: \textbf{{Self-Gain}} + Fellow Agent: \textbf{{Self-Gain}} $\Rightarrow$ Action $=$ \textbf{{Self-Gain}}
\\
\\
\textit{Note: selecting the action which maximises the \textbf{{Common-Gain}} and \textbf{{Self-Gain}} are both correct 50\% of the time over all setups.}

\end{tcolorbox}

\begin{tcolorbox}[breakable,colframe=black!75!black,title=Prisoner's Dilemma: Preferred Behaviours]
Acting Agent: \textbf{{Common-Gain}} + Fellow Agent: \textbf{{Common-Gain}} $\Rightarrow$ Action $=$ \textbf{{Common-Gain}}

Acting Agent: \textbf{{Common-Gain}} + Fellow Agent: \textbf{{Self-Gain}} $\Rightarrow$ Action $=$ \textbf{{Self-Gain}}

Acting Agent: \textbf{{Self-Gain}} + Fellow Agent: \textbf{{Common-Gain}} $\Rightarrow$ Action $=$ \textbf{{Self-Gain}}

Acting Agent: \textbf{{Self-Gain}} + Fellow Agent: \textbf{{Self-Gain}} $\Rightarrow$ Action $=$ \textbf{{Self-Gain}}
\\
\\
\textit{Note: selecting the action which maximises the \textbf{{Common-Gain}} is the correct action only 25\%, while \textbf{{Self-Gain}} is correct 75\% of the time of over all setups.}
\end{tcolorbox}

\subsection{Base Prompts}
\label{ssec:app_prompt}

\begin{table}[ht]
\caption{Payoff matrices for Stag Hunt for different configurations. The configurations differ by changing the order in which the payoff matrix is prompted to the LLM and how the payoffs are associated with the labels.}
\label{tab:configs_sh}
\begin{center}
\renewcommand{\arraystretch}{1.2}
\begin{tabular}{cccccc}
            & \multicolumn{5}{c}{Stag Hunt}                                \\ \hline
            & \multicolumn{2}{c}{Config 1}   &            & \multicolumn{2}{c}{Config 2}   \\ \hline
            & A  & B             &            & A  & B             \\ \cline{2-3} \cline{5-6} 
\multicolumn{1}{c|}{A} & 5, 5 & \multicolumn{1}{c|}{0, 4} & \multicolumn{1}{c|}{A} & 2, 2 & \multicolumn{1}{c|}{4, 0} \\
\multicolumn{1}{c|}{B} & 4, 0 & \multicolumn{1}{c|}{2, 2} & \multicolumn{1}{c|}{B} & 0, 4 & \multicolumn{1}{c|}{5, 5} \\ \cline{2-3} \cline{5-6} 
            & \multicolumn{2}{c}{Config 3}   &            & \multicolumn{2}{c}{Config 4}   \\
            & B  & A             &            & B  & A             \\ \cline{2-3} \cline{5-6} 
\multicolumn{1}{c|}{B} & 5, 5 & \multicolumn{1}{c|}{0, 4} & \multicolumn{1}{c|}{B} & 2, 2 & \multicolumn{1}{c|}{4, 0} \\
\multicolumn{1}{c|}{A} & 4, 0 & \multicolumn{1}{c|}{2, 2} & \multicolumn{1}{c|}{A} & 0, 4 & \multicolumn{1}{c|}{5, 5} \\ \cline{2-3} \cline{5-6} 
\end{tabular}
\end{center}
\end{table}

\begin{table}[ht]
\caption{Payoff matrices for Prisoner's Dilemma for the different configurations. The configurations differ by changing the order in which the payoff matrix is prompted to the LLM and how the payoffs are associated with the labels.}
\label{tab:configs_pd}

\begin{center}
\renewcommand{\arraystretch}{1.2}
\begin{tabular}{cccccc}
            & \multicolumn{5}{c}{Prisoner's Dilemma}                            \\ \hline
            & \multicolumn{2}{c}{Config 1}   &            & \multicolumn{2}{c}{Config 2}   \\ \hline
            & A  & B             &            & A  & B             \\ \cline{2-3} \cline{5-6} 
\multicolumn{1}{c|}{A} & 2, 2 & \multicolumn{1}{c|}{0, 3} & \multicolumn{1}{c|}{A} & 1, 1 & \multicolumn{1}{c|}{3, 0} \\
\multicolumn{1}{c|}{B} & 3, 0 & \multicolumn{1}{c|}{1, 1} & \multicolumn{1}{c|}{B} & 0, 3 & \multicolumn{1}{c|}{2, 2} \\ \cline{2-3} \cline{5-6} 
            & \multicolumn{2}{c}{Config 3}   &            & \multicolumn{2}{c}{Config 4}   \\
            & B  & A             &            & B  & A             \\ \cline{2-3} \cline{5-6} 
\multicolumn{1}{c|}{B} & 2, 2 & \multicolumn{1}{c|}{0, 3} & \multicolumn{1}{c|}{B} & 1, 1 & \multicolumn{1}{c|}{3, 0} \\
\multicolumn{1}{c|}{A} & 3, 0 & \multicolumn{1}{c|}{1, 1} & \multicolumn{1}{c|}{A} & 0, 3 & \multicolumn{1}{c|}{2, 2} \\ \cline{2-3} \cline{5-6} 
\end{tabular}
\end{center}
\end{table}

For each combination of preferred behaviour, different prompt configurations (payoff matrices for each config seen in Tables \ref{tab:configs_sh} and \ref{tab:configs_pd}) will be tested. The base prompt is as follows:

\begin{tcolorbox}[breakable,colframe=black!75!black,title=BASE PROMPT]
Game Context:
\\
\\
You are one player in a 2-player game. In this 2-player game, you and your fellow player balance risk and reward. Both of you make simultaneous choices between actions {label\_1} and {label\_2}. The outcomes are based on the combination of your choices. If both choose {label\_1}, you both gain PO({label\_1}, {label\_1}) point(s). Opting for {label\_1} while your opponent picks {label\_2} earns you PO({label\_1}, {label\_2}), and they get PO({label\_2}, {label\_1}). If both go for {label\_2}, you both receive PO({label\_2}, {label\_2}) point(s). Choosing {label\_2} while your opponent selects {label\_1} results in PO({label\_2}, {label\_1}) points for you and PO({label\_1}, {label\_2}) for them.
\\
\\
Player Preferences:
\\
\\
You: \texttt{{acting agent's preferred behaviour} ($CG \lor SG$)} 

Fellow Player: \texttt{{fellow agent's preferred behaviour} ($CG \lor SG$)} 
\label{prompt:base}
\end{tcolorbox}

\noindent where action labels \{{label\_1}, {label\_2}\} $= \{A, B\}$ and {label\_1} $\neq$ {label\_2}. The notation PO() refers to the payoff matrix and the corresponding gain associated with those actions being taken. Varying this prompt over the defined sets will result in a total of 16 different experimental setups, which together will provide a sufficient test suit on which the biases of the LLMs can be investigated. 

Below are the two prompting methods used: (1) Action-Only and (2) Zero-shot Chain-of-Thought prompting. 

\begin{tcolorbox}[breakable,colframe=black!75!black,title=Action-Only PROMPT]
What action should you take given the game context and player preferences?
\\
\\
You should only respond in JSON format as described below:
\\
\\
\{
\\
``action'': ``selected action, either {label\_1} or {label\_2}''
\\
\}
\label{prompt:ao}
\end{tcolorbox}

\begin{tcolorbox}[breakable,colframe=black!75!black,title=Zero-shot Chain-of-Thought PROMPT]
What action should you take given the game context and player preferences?
\\
\\
You should only respond in JSON format as described below:
\\
\\
\{
\\
``reasoning": ``step-by-step reasoning'',
\\
``action'': ``selected action, either {label\_1} or {label\_2}''
\\
\}
\\
\\
Let's work this out in a step by step way to be sure that we have the right answer.
\label{prompt:zs_cot}
\end{tcolorbox}

\section{Main Results}
\label{ssec:app_res_tabs}

For the reader's interest, the following section describes in more fine-grained detail the alignment analysis performed for all models of interest. Additionally, we provide actual responses returned by GPT-4-Turbo under different configurations. 

\subsection{Performance Analysis}

\begin{table}[ht]
\begin{center}
\renewcommand{\arraystretch}{1.2}
\begin{tabular}{cccccc}
\hline
           &              & \multicolumn{2}{c}{SH} & \multicolumn{2}{c}{PD}    \\
           &              & AO     & CoT    & AO      & CoT      \\ \hline
\multirow{2}{*}{\textbf{{GPT-3.5}}} &
 \multicolumn{1}{c|}{Con} &
 \textbf{51.8} &
 \textbf{57.2} &
 41.9 &
 42.1 \\
           & \multicolumn{1}{c|}{Self} & 48.2    & 41.6   & \textbf{58.1} & \textbf{57.1} \\ \hline
\multirow{2}{*}{\textbf{{GPT-4-Turbo}}} &
 \multicolumn{1}{c|}{Con} &
 \textbf{72.7} &
 \textbf{79.8} &
 43.3 &
 \textbf{69.2} \\
           & \multicolumn{1}{c|}{Self} & 27.3    & 20.3   & \textbf{56.8} & 30.8     \\ \hline
\multirow{2}{*}{\textbf{{GPT-4o}}} &
 \multicolumn{1}{c|}{Con} &
 \textbf{75.6} &
 \textbf{79.9} &
 \textbf{72.2} &
 \textbf{59.7} \\
           & \multicolumn{1}{c|}{Self} & 24.4    & 20.1   & 27.8     & 40.4     \\ \hline
\multicolumn{1}{l}{\multirow{2}{*}{\textbf{{Llama-3-8b}}}} &
 \multicolumn{1}{c|}{Con} &
 \textbf{50.1} &
 \textbf{60.7} &
 50.0 &
 43.2 \\
\multicolumn{1}{l}{} & \multicolumn{1}{c|}{Self} & 49.9    & 38.9   & 50.0     & \textbf{56.6} \\ \hline
\end{tabular}
\end{center}
\caption{Table showing frequency at which the considerate (Con) and the selfish (Self) actions are selected. Results are shown for both games, Stag Hunt (SH) and Prisoner's Dilemma (PD). Each game is tested using Answer-Only (AO) and Zero-shot Chain-of-Thought (CoT) prompt.}
\label{tab:cons_self_summary}
\end{table}

We can see in Figure \ref{fig:acc} a more detailed alignment analysis, where the performance comparison is performed for each bias, under each prompting scheme for all models. Additionally, in Figure \ref{fig:sig_vs_diff} a plot showing the relationship between the observed performance drop in the LLMs when misaligned and the statistical significance of the biases can be seen. We found that for both games, there is a positive statistically significant relation between the two, meaning that the more significant the bias the larger the drop in accuracy observed. 

\begin{figure*}[ht]
\begin{center}
\includegraphics[width=0.45\textwidth]{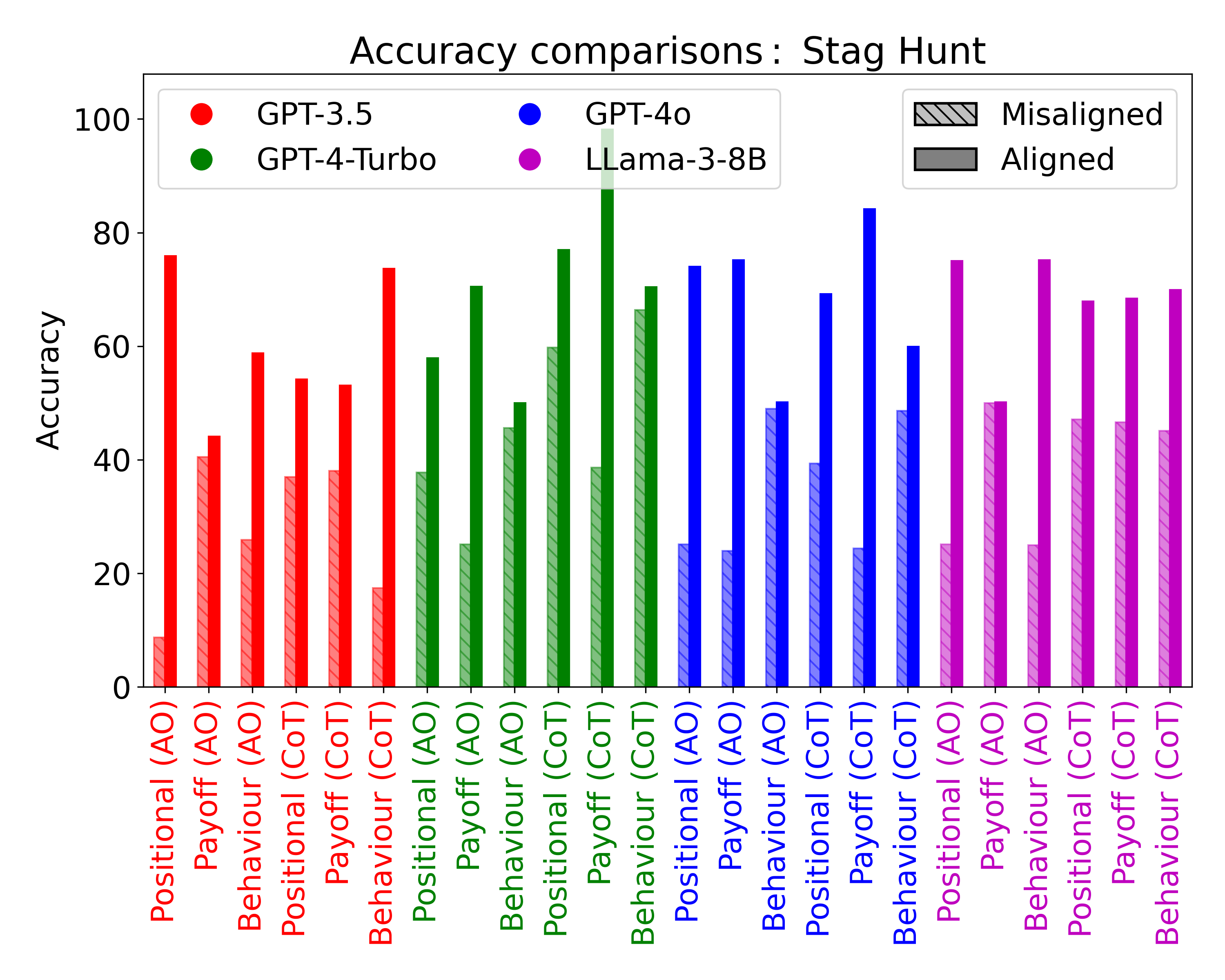} \hfill
\includegraphics[width=0.45\textwidth]{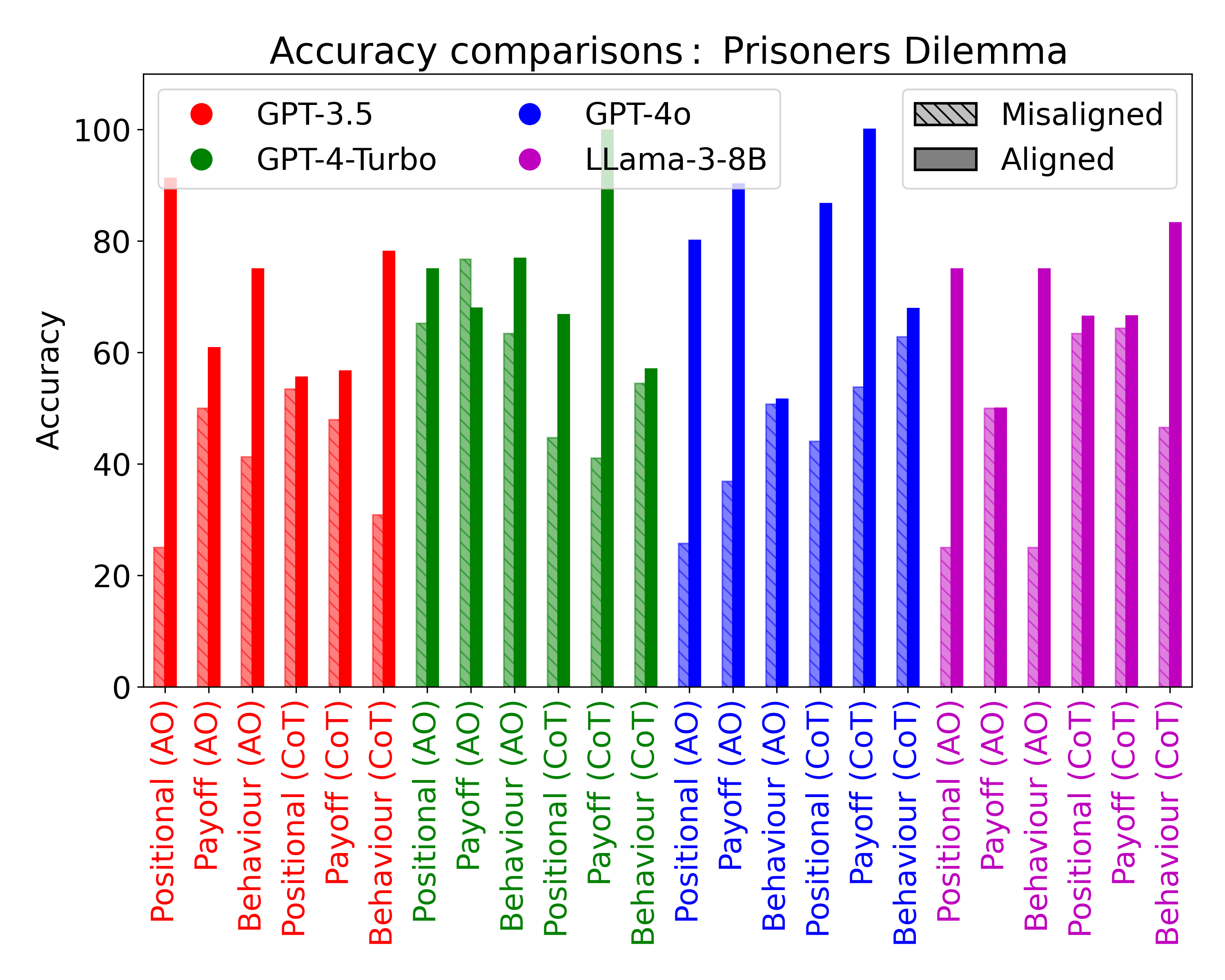} \hfill
\caption {Figure showing the alignment analysis for all models across all systematic biases. We see the comparison in the performance (measured based on the selection of the correct action given the prompted preferred behaviours) for each model under the two tested prompting methods: (1) Answer-Only (AO) and (2) Chain-of-Thought (CoT). It is clear that for almost all configurations, when the bias is misaligned the performance suffers greatly.}
\label{fig:acc}
\end{center}
\end{figure*}

\begin{figure*}[ht]
\begin{center}
\includegraphics[width=0.45\textwidth]{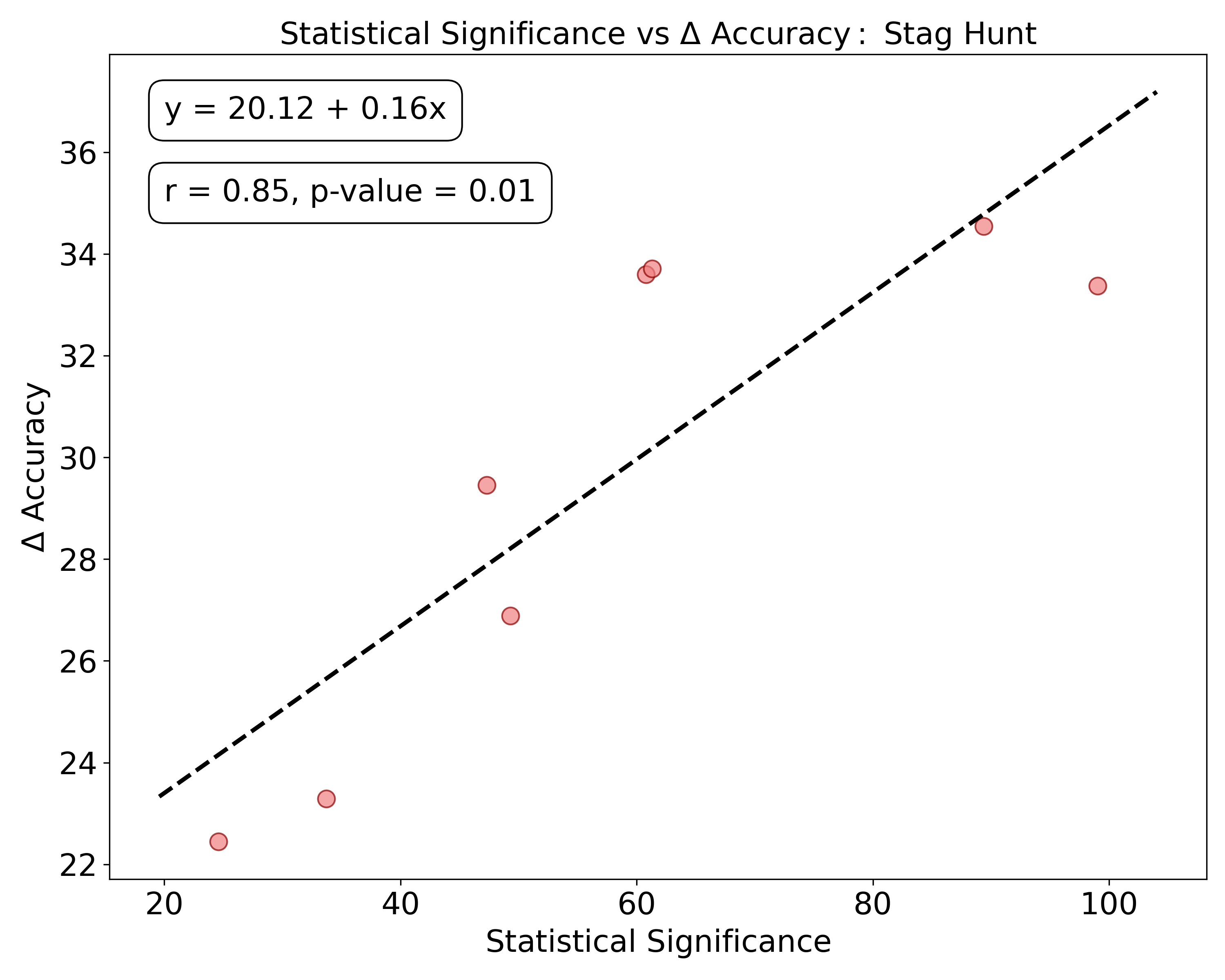} \hfill
\includegraphics[width=0.45\textwidth]{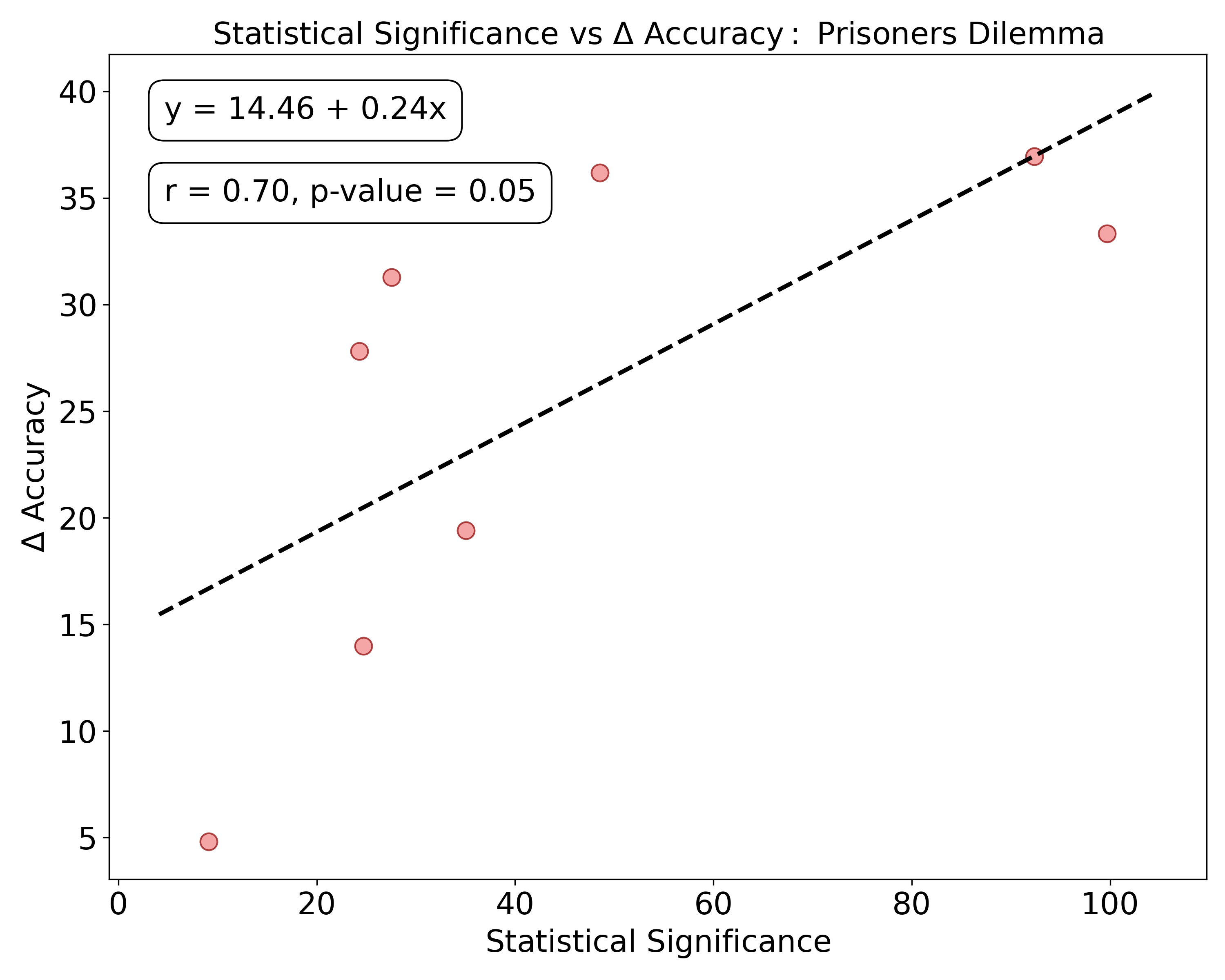} \hfill
\caption {Figure showing average statistical significance vs the difference in accuracy achieved when misaligned and aligned. (LEFT) Stag Hunt - the correlation coefficient calculated, $r=0.85$, is shown to be statistically significant with a $\rho=0.01$ and (RIGHT) Prisoner Dilemma - the correlation coefficient calculated, $r=0.70$, is shown to be statistically significant with a $\rho=0.05$. We can, therefore, say that the linear relationship between the average statistical significance of the three measured biases and the performance drop when the experiment setting is misaligned is statistically significant in both games.}
\label{fig:sig_vs_diff}
\end{center}
\end{figure*}

\subsection{LLM Reasoning}

The following blocks of text show common patterns of reasoning for GPT4-Turbo under the 4 different experimental configurations when both agents are prompted with a preference to prioritise their self-gain (similar patterns are observed for other combinations of preferred behaviours). Note how the LLM uses reasoning over the action it will select changes as the configurations change. This is because it is biased towards selecting a certain action and will therefore reason in such a way that the biased answer might make sense.

\begin{tcolorbox}[breakable, colframe=black!75!black,title=Stag Hunt: \\ Changes in reasoning across configs]
\begin{tcolorbox}[colframe=black!75!black,title=Configuration 1]
Given the game's rules and the principles of both players, we need to consider the outcomes based on mutual self-interest. If both players choose A, each gains 5 points, which is the maximum possible for mutual choice. If one chooses A and the other B, the player choosing B gains 4 points while the other gains 0. If both choose B, each gains 2 points. Since both players prioritize their own points, choosing A could risk gaining 0 points if the other player chooses B, aiming for 4 points. However, choosing B guarantees at least 2 points, and potentially 4 if the other player chooses A. Given the risk of gaining 0 points with A and the guarantee of at least 2 points with B, the safer and potentially more rewarding choice is B.
\end{tcolorbox}
\begin{tcolorbox}[colframe=black!75!black,title=Configuration 2]
Given the game's rules and the principles of both players, we need to consider the outcomes based on mutual decisions. If both choose A, each gains 2 points. If one chooses A and the other B, the one who chooses A gets 4 points, and the one who chooses B gets 0. If both choose B, each gains 5 points. If one chooses B and the other A, the one who chooses B gets 0 points, and the one who chooses A gets 4. Since both players prioritize their own points, choosing B is the safer and potentially more rewarding option because the worst-case scenario (choosing B while the opponent chooses A) results in 0 points, but the best-case scenario (both choosing B) results in 5 points each, which is the maximum possible gain for both players simultaneously.
\end{tcolorbox}
\begin{tcolorbox}[colframe=black!75!black,title=Configuration 3]
Given the game's rules and the principles of both players, we can analyze the outcomes. If both players choose B, each gains 5 points, which is a good outcome. If one chooses B and the other A, the one choosing A gains 4 points while the other gets 0. If both choose A, each gains 2 points. Since both players prioritize their own points, choosing B could risk getting 0 points if the other player chooses A, aiming for 4 points. However, choosing A guarantees at least 2 points regardless of the other player's choice, and potentially 4 points if the other chooses B. Therefore, choosing A maximizes the guaranteed points and also offers a chance for a higher reward if the opponent chooses B.
\end{tcolorbox}
\begin{tcolorbox}[colframe=black!75!black,title=Configuration 4]
Given the game's rules and the principles of both players, we need to consider the outcomes based on each combination of choices. If both players choose B, each earns 2 points. If one chooses B and the other A, the B chooser gets 4 points and the A chooser gets 0. If both choose A, each earns 5 points. Since both players prioritize their own points, choosing A could be risky if the other player chooses B, resulting in 0 points. However, choosing B is safer as the worst outcome is 2 points (if both choose B) and the best is 4 points (if the opponent chooses A). Given that both players aim to maximize their own points, it's likely the opponent might also consider choosing B to avoid the risk of earning 0 points. Therefore, choosing B seems to be the safer and potentially more rewarding strategy based on the game setup and player principles.
\end{tcolorbox}
\end{tcolorbox}

\begin{tcolorbox}[breakable, colframe=black!75!black,title=Prisoner's Dilemma: \\ Changes in reasoning across configs]
\begin{tcolorbox}[colframe=black!75!black,title=Configuration 1]
Given the game's rules and the principles of both players, we need to consider the outcomes of each combination of choices. If both choose A, each gains 2 points. If one chooses A and the other B, the one who chooses B gains 3 points while the other gains 0. If both choose B, each gains 1 point. Since both players prioritize their own points, choosing B is a safer option because it guarantees at least 1 point and potentially 3 points if the opponent chooses A. Choosing A risks gaining 0 points if the opponent chooses B.
\end{tcolorbox}
\begin{tcolorbox}[colframe=black!75!black,title=Configuration 2]
Given the game's rules and the principles of both players, we need to consider the outcomes of each combination of choices. If both players choose A, each gains 1 point. If one chooses A and the other B, the one who chooses A gains 3 points while the other gets 0. If both choose B, each gains 2 points. If one chooses B and the other A, the one who chooses B gets 0 points and the other gets 3. Since both players prioritize their own points, choosing B is safer as it guarantees at least 2 points if both choose B, whereas choosing A risks getting 0 points if the opponent chooses B. Therefore, choosing B maximizes the minimum gain (minimax strategy).
\end{tcolorbox}
\begin{tcolorbox}[colframe=black!75!black,title=Configuration 3]
Given that both players prioritize their own points, choosing action B is risky unless both players choose B, which results in 2 points each. However, if one chooses B and the other chooses A, the player who chose A gets 3 points while the one who chose B gets 0. If both choose A, each player gets 1 point. Since both players aim to maximize their own points, it\'s likely that the opponent will choose A to avoid the risk of getting 0 points by choosing B. Therefore, to maximize your own points in anticipation of the opponent\'s likely choice of A, you should also choose A to ensure you get at least 1 point rather than risking getting 0 points.
\end{tcolorbox}
\begin{tcolorbox}[colframe=black!75!black,title=Configuration 4]
Given the game's rules and the principles of both players, we need to consider the outcomes based on each possible combination of choices. If both players choose B, each earns 1 point. If one chooses B and the other A, the player choosing B earns 3 points while the other earns 0. If both choose A, each earns 2 points. If one chooses A and the other B, the player choosing A earns 0 points while the other earns 3. Since both players prioritize their own points, choosing A could risk earning 0 points if the other player chooses B. However, choosing B guarantees at least 1 point and can earn up to 3 points if the other player chooses A. Therefore, choosing B maximizes the potential gain while minimizing the risk of earning no points.
\end{tcolorbox}
\end{tcolorbox}

\section{Additional Results}
\label{sec:add_results}

For the reader's interest, the following section provides further detail on the additional experiments and results performed, specifically: (i) exploring the impact of fine-tuning on these biases, (ii) testing how prior knowledge affects LLM biases, (iii) evaluating how different prompt structures influence observed biases, and (iv) comparing the behaviours of humans and LLMs in a two-player zero-sum game. 

\begin{table}[ht!]
\begin{center}
\renewcommand{\arraystretch}{1.2}
\begin{tabular}{ccccccc}
\hline
 &
  &
  &
 \multicolumn{2}{c}{SH} &
 \multicolumn{2}{c}{PD} \\
 &
  &
  &
 AO &
 CoT &
 AO &
 CoT \\ \hline
\multirow{4}{*}{\textbf{Base}} &
 \multirow{2}{*}{SG} &
 \multicolumn{1}{c|}{A} &
 9.9 &
 \multicolumn{1}{c|}{6.1} &
 10.00 &
 9.00 \\
 &
  &
 \multicolumn{1}{c|}{B} &
 \textbf{10.1} &
 \multicolumn{1}{c|}{\textbf{13.9}} &
 10.00 &
 \textbf{11.00} \\ \cline{4-7} 
            & \multirow{2}{*}{CG} & \multicolumn{1}{c|}{A} & \textbf{10.4} & \multicolumn{1}{c|}{\textbf{10.8}} & \textbf{12.13} & \textbf{11.00} \\
 &
  &
 \multicolumn{1}{c|}{B} &
 9.6 &
 \multicolumn{1}{c|}{9.3} &
 7.88 &
 9.00 \\ \hline
\multirow{4}{*}{\textbf{Leaked}} & \multirow{2}{*}{SG} & \multicolumn{1}{c|}{A} & 10.0     & \multicolumn{1}{c|}{7.9}      & 10.00     & 10.00     \\
 &
  &
 \multicolumn{1}{c|}{B} &
 10.0 &
 \multicolumn{1}{c|}{\textbf{12.1}} &
 10.00 &
 10.00 \\ \cline{4-7} 
 &
 \multirow{2}{*}{CG} &
 \multicolumn{1}{c|}{A} &
 \textbf{12.1} &
 \multicolumn{1}{c|}{8.5} &
 10.00 &
 9.38 \\
 &
  &
 \multicolumn{1}{c|}{B} &
 7.9 &
 \multicolumn{1}{c|}{\textbf{11.5}} &
 10.00 &
 \textbf{10.63} \\ \hline
\multirow{4}{*}{\textbf{Bullet}} &
 \multirow{2}{*}{SG} &
 \multicolumn{1}{c|}{A} &
 5.6 &
 \multicolumn{1}{c|}{8.9} &
 7.50 &
 6.38 \\
 &
  &
 \multicolumn{1}{c|}{B} &
 \textbf{14.4} &
 \multicolumn{1}{c|}{\textbf{11.1}} &
 \textbf{12.50} &
 \textbf{13.63} \\ \cline{4-7} 
 &
 \multirow{2}{*}{CG} &
 \multicolumn{1}{c|}{A} &
 \textbf{10.0} &
 \multicolumn{1}{c|}{\textbf{10.3}} &
 10.00 &
 9.13 \\
 &
  &
 \multicolumn{1}{c|}{B} &
 10.0 &
 \multicolumn{1}{c|}{9.8} &
 10.00 &
 \textbf{10.88} \\ \hline
\multicolumn{1}{l}{\multirow{4}{*}{\textbf{Code}}} &
 SG &
 \multicolumn{1}{c|}{A} &
 7.1 &
 \multicolumn{1}{c|}{9.8} &
 8.88 &
 8.25 \\
\multicolumn{1}{l}{}  &           & \multicolumn{1}{c|}{B} & \textbf{12.9} & \multicolumn{1}{c|}{\textbf{10.3}} & \textbf{11.13} & \textbf{11.75} \\ \cline{4-7} 
\multicolumn{1}{l}{}  & \multirow{2}{*}{CG} & \multicolumn{1}{c|}{A} & \textbf{12.6} & \multicolumn{1}{c|}{9.9}      & \textbf{16.00} & \textbf{10.13} \\
\multicolumn{1}{l}{} &
  &
 \multicolumn{1}{c|}{B} &
 7.4 &
 \multicolumn{1}{c|}{\textbf{10.1}} &
 4.00 &
 9.88 \\ \hline
\end{tabular}
\end{center}
\caption{Table showing frequency at which labels A and B are selected based on the prompted behaviours, Self-Gain (SG) and Common-Gain (CG), of the LLM. Results are shown for both games, Stag Hunt (SH) and Prisoner's Dilemma (PD). Each game is tested using Answer-Only (AO) and Zero-shot Chain-of-Thought (CoT) prompting.}
\label{tab:add_res_behave}
\end{table}

\begin{table}[ht!]
\begin{center}
\renewcommand{\arraystretch}{1.2}
\begin{tabular}{cccccc}
\hline
                 &              & \multicolumn{2}{c}{SH}     & \multicolumn{2}{c}{PD}     \\
                 &              & AO       & CoT      & AO       & CoT      \\ \hline
\multirow{2}{*}{\textbf{Base}}  & \multicolumn{1}{c|}{Con} & \textbf{14.88} & \textbf{15.69} & \textbf{13.94} & \textbf{12.50} \\
                 & \multicolumn{1}{c|}{Self} & 5.13      & 4.31      & 6.06      & 7.50      \\ \hline
\multirow{2}{*}{\textbf{Leaked}} & \multicolumn{1}{c|}{Con} & \textbf{10.19} & \textbf{11.44} & 10.00     & 5.94      \\
                 & \multicolumn{1}{c|}{Self} & 9.81      & 8.56      & 10.00     & \textbf{14.06} \\ \hline
\multirow{2}{*}{\textbf{Bullet}} & \multicolumn{1}{c|}{Con} & \textbf{17.31} & \textbf{17.69} & \textbf{13.75} & \textbf{12.75} \\
                 & \multicolumn{1}{c|}{Self} & 2.69      & 2.31      & 6.25      & 7.25      \\ \hline
\multirow{2}{*}{\textbf{Code}}  & \multicolumn{1}{c|}{Con} & \textbf{12.13} & \textbf{12.69} & 7.56      & 8.81      \\
                 & \multicolumn{1}{c|}{Self} & 7.88      & 7.31      & \textbf{12.44} & \textbf{11.19} \\ \hline
\end{tabular}
\end{center}
\caption{Table showing frequency at which the considerate (Con) and the selfish (Self) actions are selected. Results are shown for both games, Stag Hunt (SH) and Prisoner's Dilemma (PD). Each game is tested using Answer-Only (AO) and Zero-shot Chain-of-Thought (CoT) prompting.}
\label{tab:add_res_cons_self_summary}
\end{table}

\begin{table*}[ht]
\begin{center}
\renewcommand{\arraystretch}{1.2}
\begin{tabular}{cccccccccc}
\hline
 &            & \multicolumn{4}{c}{SH}                       & \multicolumn{4}{c}{PD}                  \\
 &            & \multicolumn{2}{c}{AO} & \multicolumn{2}{c}{CoT}          & \multicolumn{2}{c}{AO} & \multicolumn{2}{c}{CoT}     \\
 &
  &
 A $\rightarrow$ B &
 B $\rightarrow$ A &
 A $\rightarrow$ B &
 B $\rightarrow$ A &
 A $\rightarrow$ B &
 B $\rightarrow$ A &
 A $\rightarrow$ B &
 B $\rightarrow$ A \\ \hline
\multirow{2}{*}{\textbf{Base}} &
 \multicolumn{1}{c|}{A} &
 5.00 &
 \textbf{15.25} &
 6.00 &
 \multicolumn{1}{c|}{\textbf{10.88}} &
 7.13 &
 \textbf{15.00} &
 6.13 &
 \textbf{13.88} \\
 & \multicolumn{1}{c|}{B} & \textbf{15.00} & 4.75 & \textbf{14.00} & \multicolumn{1}{c|}{9.13} & \textbf{12.88} & 5.00 & \textbf{13.88} & 6.13      \\ \hline
\multirow{2}{*}{\textbf{Leaked}} &
 \multicolumn{1}{c|}{A} &
 8.50 &
 \textbf{13.63} &
 5.13 &
 \multicolumn{1}{c|}{\textbf{11.25}} &
 10.00 &
 10.00 &
 9.50 &
 9.88 \\
 & \multicolumn{1}{c|}{B} & \textbf{11.50} & 6.38 & \textbf{14.88} & \multicolumn{1}{c|}{8.75} & 10.00     & 10.00 & \textbf{10.50} & \textbf{10.13} \\ \hline
\multirow{2}{*}{\textbf{Bullet}} &
 \multicolumn{1}{c|}{A} &
 5.13 &
 \textbf{10.50} &
 7.25 &
 \multicolumn{1}{c|}{\textbf{11.88}} &
 5.00 &
 \textbf{12.50} &
 4.00 &
 \textbf{11.50} \\
 & \multicolumn{1}{c|}{B} & \textbf{14.88} & 9.50 & \textbf{12.75} & \multicolumn{1}{c|}{8.13} & \textbf{15.00} & 7.50 & \textbf{16.00} & 8.50      \\ \hline
\multirow{2}{*}{\textbf{Code}} &
 \multicolumn{1}{c|}{A} &
 4.25 &
 \textbf{15.38} &
 5.13 &
 \multicolumn{1}{c|}{\textbf{14.63}} &
 \textbf{13.25} &
 \textbf{11.63} &
 6.00 &
 \textbf{12.38} \\
 & \multicolumn{1}{c|}{B} & \textbf{15.75} & 4.63 & \textbf{14.88} & \multicolumn{1}{c|}{5.38} & 6.75      & 8.38 & \textbf{14.00} & 7.63      \\ \hline
\end{tabular}
\end{center}
\caption{Table showing frequency at which labels A and B are selected based on the prompted order (A $\rightarrow$ B or B $\rightarrow$ A) of the LLM. Results are shown for both games, Stag Hunt (SH) and Prisoner's Dilemma (PD). Each game is tested using Answer-Only (AO) and Zero-shot Chain-of-Thought (CoT) prompting.}
\label{tab:add_res_order_summary}
\end{table*}

\begin{table*}[ht]
\begin{center}
\renewcommand{\arraystretch}{1.2}
\renewcommand{\arraystretch}{1.2}
\begin{tabular}{cccccccccc}
\hline
 &            & \multicolumn{4}{c}{SH}                        & \multicolumn{4}{c}{PD}                  \\
 &            & \multicolumn{2}{c}{AO} & \multicolumn{2}{c}{CoT}           & \multicolumn{2}{c}{AO} & \multicolumn{2}{c}{CoT}     \\
 &            & A=C  & B=C      & A=C  & B=C                 & A=C  & B=C      & A=C      & B=C      \\ \hline
\multirow{2}{*}{\textbf{Base}} &
 \multicolumn{1}{c|}{A} &
 \textbf{15.00} &
 8.20 &
 \textbf{14.13} &
 \multicolumn{1}{c|}{2.75} &
 \textbf{15.00} &
 9.70 &
 \textbf{12.50} &
 7.50 \\
 & \multicolumn{1}{c|}{B} & 5.00 & \textbf{11.80} & 5.88 & \multicolumn{1}{c|}{\textbf{17.25}} & 5.00 & \textbf{10.30} & 7.50      & \textbf{12.50} \\ \hline
\multirow{2}{*}{\textbf{Leaked}} &
 \multicolumn{1}{c|}{A} &
 \textbf{11.25} &
 10.70 &
 \textbf{9.63} &
 \multicolumn{1}{c|}{6.75} &
 10.00 &
 10.00 &
 5.63 &
 \textbf{13.75} \\
 & \multicolumn{1}{c|}{B} & 8.75 & \textbf{9.30} & 10.38 & \multicolumn{1}{c|}{\textbf{13.25}} & 10.00 & 10.00     & \textbf{14.38} & 6.25      \\ \hline
\multirow{2}{*}{\textbf{Bullet}} &
 \multicolumn{1}{c|}{A} &
 \textbf{15.13} &
 4.40 &
 \textbf{17.25} &
 \multicolumn{1}{c|}{1.88} &
 \textbf{12.50} &
 8.00 &
 \textbf{10.50} &
 5.00 \\
 & \multicolumn{1}{c|}{B} & 4.88 & \textbf{15.60} & 2.75 & \multicolumn{1}{c|}{\textbf{18.13}} & 7.50 & \textbf{12.00} & 9.50      & \textbf{15.00} \\ \hline
\multirow{2}{*}{\textbf{Code}} &
 \multicolumn{1}{c|}{A} &
 \textbf{12.00} &
 9.70 &
 \textbf{12.50} &
 \multicolumn{1}{c|}{7.13} &
 10.00 &
 \textbf{13.90} &
 8.00 &
 \textbf{10.38} \\
 & \multicolumn{1}{c|}{B} & 8.00 & \textbf{10.30} & 7.50 & \multicolumn{1}{c|}{\textbf{12.88}} & 10.00 & 6.10      & \textbf{12.00} & 9.63      \\ \hline
\end{tabular}
\end{center}
\caption{Table showing the frequency at which labels A and B are selected based on the prompted order (A=C or B=C) of the LLM. Results are shown for both games, Stag Hunt (SH) and Prisoner's Dilemma (PD). Each game is tested using Answer-Only (AO) and Zero-shot Chain-of-Thought (CoT) prompting.}
\label{tab:add_res_payoff_summary}
\end{table*}

\begin{figure*}[ht]
\includegraphics[width=0.45\textwidth]{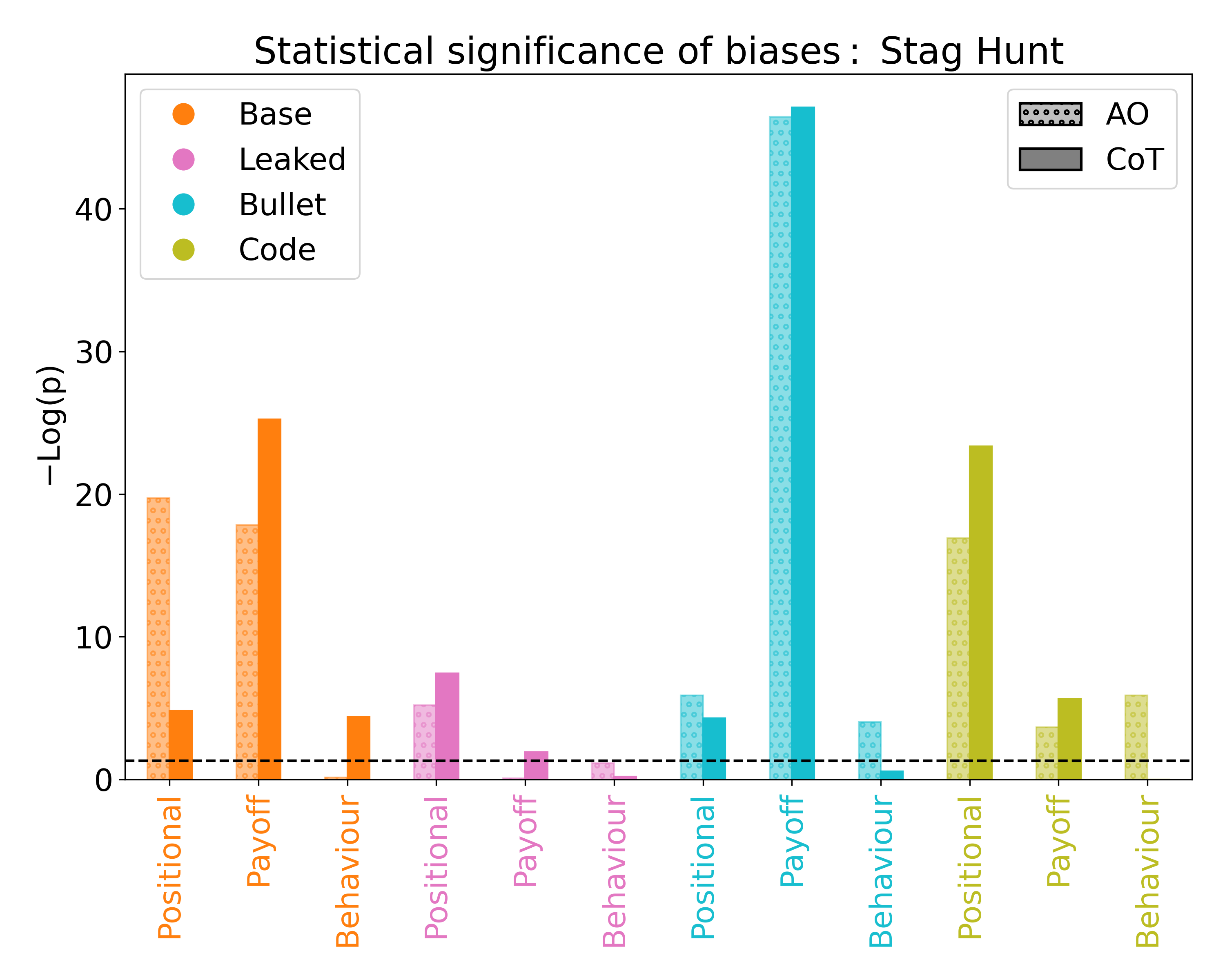}
\includegraphics[width=0.45\textwidth]{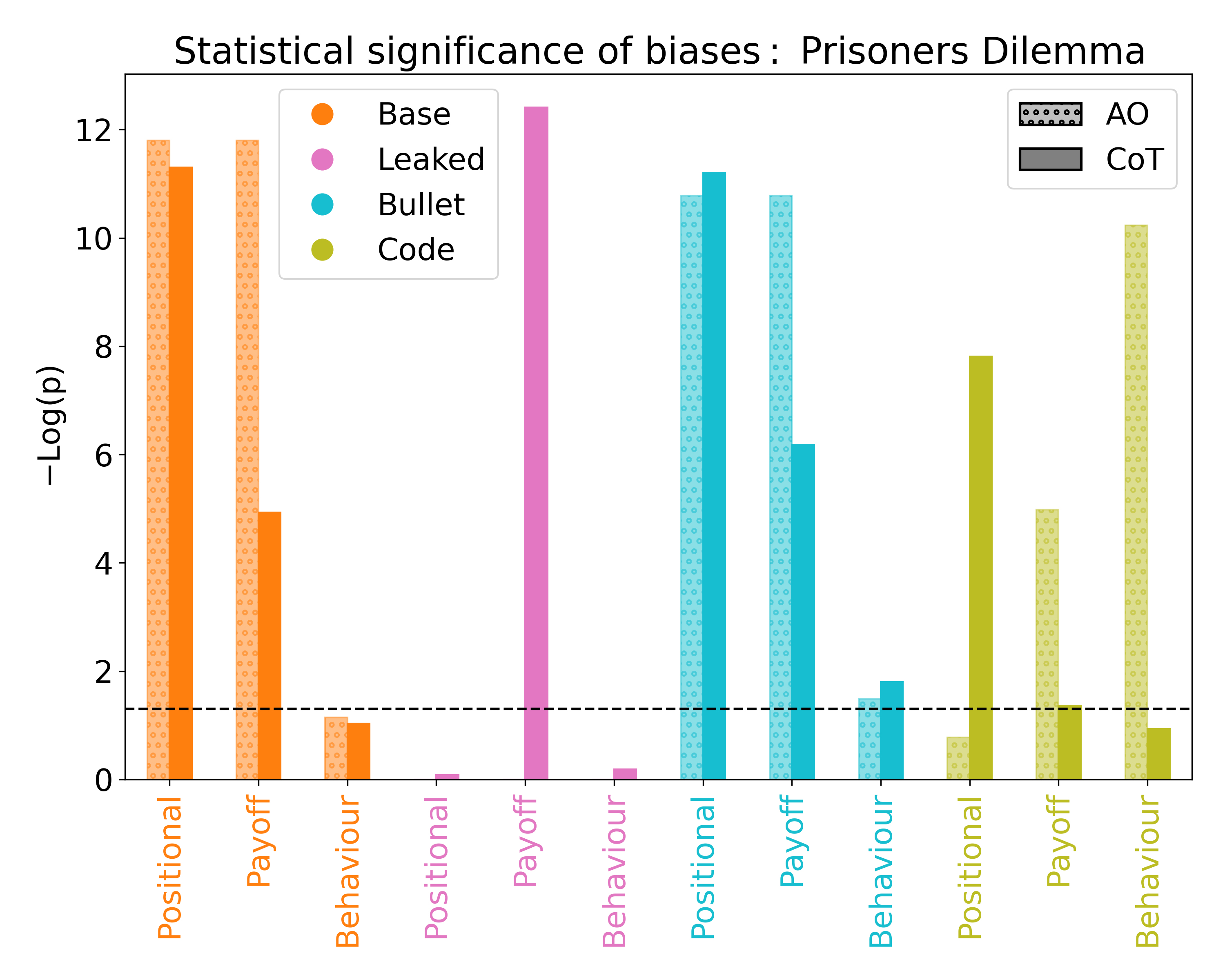}
 \caption{Figure showing the statistical analysis of the identified biases for all prompting structures. The larger the $-Log(p)$, the more statistically significant the bias. The dashed \textbf{black} line signifies the threshold at which the bias becomes statistically significant (found close to the horizontal axis for both plots). Notably, it can be seen that the leaked prompt, on average, is the least affected by the biases. However, all prompting methods still result in a certain level of bias.}
 \label{fig:add_res_sig}
\end{figure*}

\subsection{Effect of Prior Knowledge}
\label{ssec:app_eff_prior}

To investigate the effect an LLM's prior knowledge has on the identified biases, we alter the BASE PROMPT (in Section~\ref{prompt:base} of the Technical Appendix) used in the main set of experiments to include the names of the games i.e. Prisoner's Dilemama and Stag Hunt. This can seen in LEAKED PROMPT. We run the same experiment setup described in Section~\ref{sec:meth}, except we run it fewer times (20x). Looking at Tables \ref{tab:add_res_behave}, \ref{tab:add_res_cons_self_summary}, \ref{tab:add_res_order_summary}, and \ref{tab:add_res_payoff_summary}, and Figures \ref{fig:add_res_acc} and \ref{fig:add_res_sig}, we found that when prompting GPT-4o with the names of the games in the prompt, the effect of the biases weakens. This drop is likely due to GPT-4o relying more on its prior knowledge about these games, rather than reasoning from scratch, to make its decisions. Since GPT-4o is likely to have encountered these games and their solutions during training, it can solve the different instances with a higher degree of accuracy.

\begin{tcolorbox}[breakable,colframe=black!75!black,title=LEAKED PROMPT]
Game Context:
\\
\\
You are one player in a 2-player game of {GAME\_NAME}. You and your fellow player balance risk and reward. Both of you make simultaneous choices between actions {label\_1} and {label\_2}. The outcomes are based on the combination of your choices. If both choose {label\_1}, you both gain PO({label\_1}, {label\_1}) point(s). Opting for {label\_1} while your opponent picks {label\_2} earns you PO({label\_1}, {label\_2}), and they get PO({label\_2}, {label\_1}). If both go for {label\_2}, you both receive PO({label\_2}, {label\_2}) point(s). Choosing {label\_2} while your opponent selects {label\_1} results in PO({label\_2}, {label\_1}) points for you and PO({label\_1}, {label\_2}) for them.
\\
\\
Player Preferences:
\\
\\
You: \texttt{{acting agent's preferred behaviour} ($CG \lor SG$)} 
Fellow Player: \texttt{{fellow agent's preferred behaviour} ($CG \lor SG$)} 
\label{prompt:leaked}
\end{tcolorbox}

\subsection{Effect of Prompt Variations}
\label{ssec:app_eff_prompt_var}

We aimed to show that these biases, which present themselves in the main set of experiments, are not specific to the base prompt used and are instead inherent in the LLM itself. We do this by testing GPT-4o under different prompt structures, which we refer to as (1) BULLET PROMPT and (2) CODE PROMPT. We run the same experiment setup described in Section~\ref{sec:meth}, except we run it fewer times (20x). Looking at Tables \ref{tab:add_res_behave}, \ref{tab:add_res_cons_self_summary}, \ref{tab:add_res_order_summary}, and \ref{tab:add_res_payoff_summary}, and Figures \ref{fig:add_res_acc} and \ref{fig:add_res_sig}, we found that the biases present themselves under all tested prompting formats, albeit not equally. This shows that by changing the prompt structure, one can alter the effect the biases have on the LLM's decision-making abilities, however, it also shows that one cannot simply expect to rid the LLM of its biases with these changes.

\begin{tcolorbox}[breakable,colframe=black!75!black,title=BULLET PROMPT]
You are one player in a 2-player game. In this 2-player game, you and your fellow player navigate a balance risk and reward.
\\
\\
**Game Setup:**
\begin{itemize}
  \item Players: 2
  \item Actions available to both players: {label\_1} and {label\_2}
  \item Players make their choices simultaneously.
\end{itemize}
**Payoff Matrix:**
\begin{itemize}
  \item Both choose {label\_1}:
  \begin{itemize}
    \item You receive: PO({label\_1}, {label\_1}) points
    \item Fellow player receives: PO({label\_1}, {label\_1}) points
  \end{itemize}
  \item You choose {label\_1}, Fellow player chooses {label\_2}::
  \begin{itemize}
    \item You receive: PO({label\_1}, {label\_2}) points
    \item Fellow player receives: PO({label\_2}, {label\_1}) points
  \end{itemize}
  \item You choose {label\_2}, Fellow player chooses {label\_1}:
  \begin{itemize}
    \item You receive: PO({label\_2}, {label\_1}) points
    \item Fellow player receives: PO({label\_1}, {label\_2}) points
  \end{itemize}
  \item Both choose {label\_2}:
  \begin{itemize}
    \item You receive: PO({label\_2}, {label\_2}) points
    \item Fellow player receives: PO({label\_2}, {label\_2}) points
  \end{itemize}
\end{itemize}
**Player Principles:**
\begin{itemize}
  \item You: \texttt{{acting agent's preferred behaviour} ($CG \lor SG$)}
  \item Fellow Player: \texttt{{fellow agent's preferred behaviour} ($CG \lor SG$)}
\end{itemize}
\label{prompt:bullet}
\end{tcolorbox}

\begin{tcolorbox}[breakable,colframe=black!75!black,title=CODE PROMPT]
You are one player in a 2-player game. In this 2-player game, you and your fellow player navigate a balance risk and reward. Both of you make simultaneous choices between actions {label\_1} and {label\_2}.
\\
\\
Game Scenario:
\vspace{5 mm} 
\begin{lstlisting}
def game_scenario(player_choice, fellow_player\_choice):
  """
  Simulates the outcome of a 2-player game based on the choices made by the players and the payoff matrix.
  
  Parameters:
  player_choice (str): The choice made by the player (either {label_1} or {label_2}).
  fellow_player_choice (str): The choice made by the fellow player (either {label_1} or {label_2}).
  
  Returns:
  dict: A dictionary containing the points earned by both the player and the fellow player.
  """

  if player_choice == {label_1} and fellow_player_choice == {label_1}:
    player_points = fellow_player_points = PO({label_1}, {label_1})
  elif player_choice == {label_1} and fellow_player_choice == {label_2}:
    player_points = PO({label_1}, {label_2})
    fellow_player_points = PO({label_2}, {label_1})
  elif player_choice == {label_2} and fellow_player_choice == {label_1}:
    player_points = PO({label_2}, {label_1})
    fellow_player_points = PO({label_1}, {label_2})
  else: # Both choose {label_2}
    player_points = fellow_player_points = PO({label_2}, {label_2})
  
  return {{
    "player_points": player_points,
    "fellow_player_points": fellow_player_points
  }}
\end{lstlisting}
\vspace{5 mm} 
Player Principles:
You: \texttt{{acting agent's preferred behaviour} ($CG \lor SG$)}
Fellow Player: \texttt{{fellow agent's preferred behaviour} ($CG \lor SG$)}
\label{prompt:code}
\end{tcolorbox}

\subsection{Effect of Fine-Tuning}
\label{ssec:app_eff_fine_tune}

We also run a set of experiments using LLama-3-8b (w/o Fine-Tuning) and LLama-3-8b-Instruct (w/ Fine-Tuning) to investigate the effect instruction fine-tuning has on the identified biases. We run the same experiment setup described in Section~\ref{sec:meth}, except we run it fewer times (20x). Looking at Figures \ref{fig:add_res_ft_acc} and \ref{fig:add_res_ft_stat}, we find that LLama-3-8b is primarily guided by the order in which the action labels are presented in the prompt, whereas LLama-3-8b-Instruct is equally guided by both the order and the preferred behaviours. However, the overall effect of the biases is weakened when the LLM is fine-tuned. This suggests that fine-tuning LLMs may have a positive effect on the model's ability to overcome these systematic biases, however, it may also affect the models in an unforeseen manner, such as introducing a behavioural bias in LLama-3-8b.


\begin{figure}[ht!]
\centering
\includegraphics[width=0.45\textwidth]{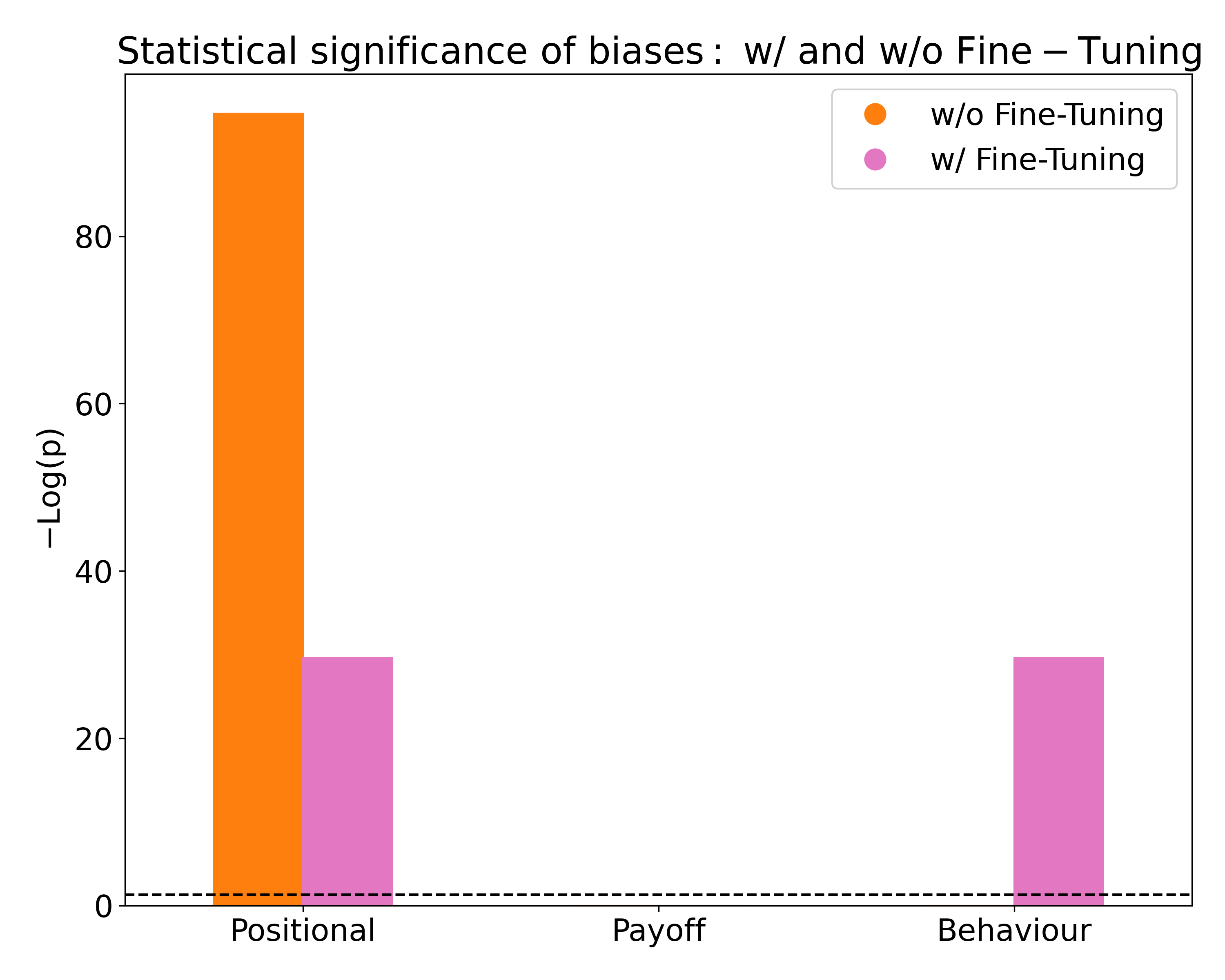}\hfill
 \caption{Figure showing the statistical analysis of the identified biases for \textbf{Llama-3-8B} w/ and w/o Fine-Tuning. The larger the $-Log(p)$, the more statistically significant the bias. The dashed \textbf{black} line signifies the threshold at which the bias becomes statistically significant (found close to the horizontal axis for both plots). Notably, it can be seen that w/o Fine-Tuning the model is only significantly affected by the positional bias, whereas w/ Fine-Tuning it is significantly affected by both position and behaviour biases.}
 \label{fig:add_res_ft_stat}
\end{figure}

\subsection{Humans vs AI}
\label{ssec:app_humvsai}

\begin{figure}[ht!]
\centering
\includegraphics[width=0.45\textwidth]{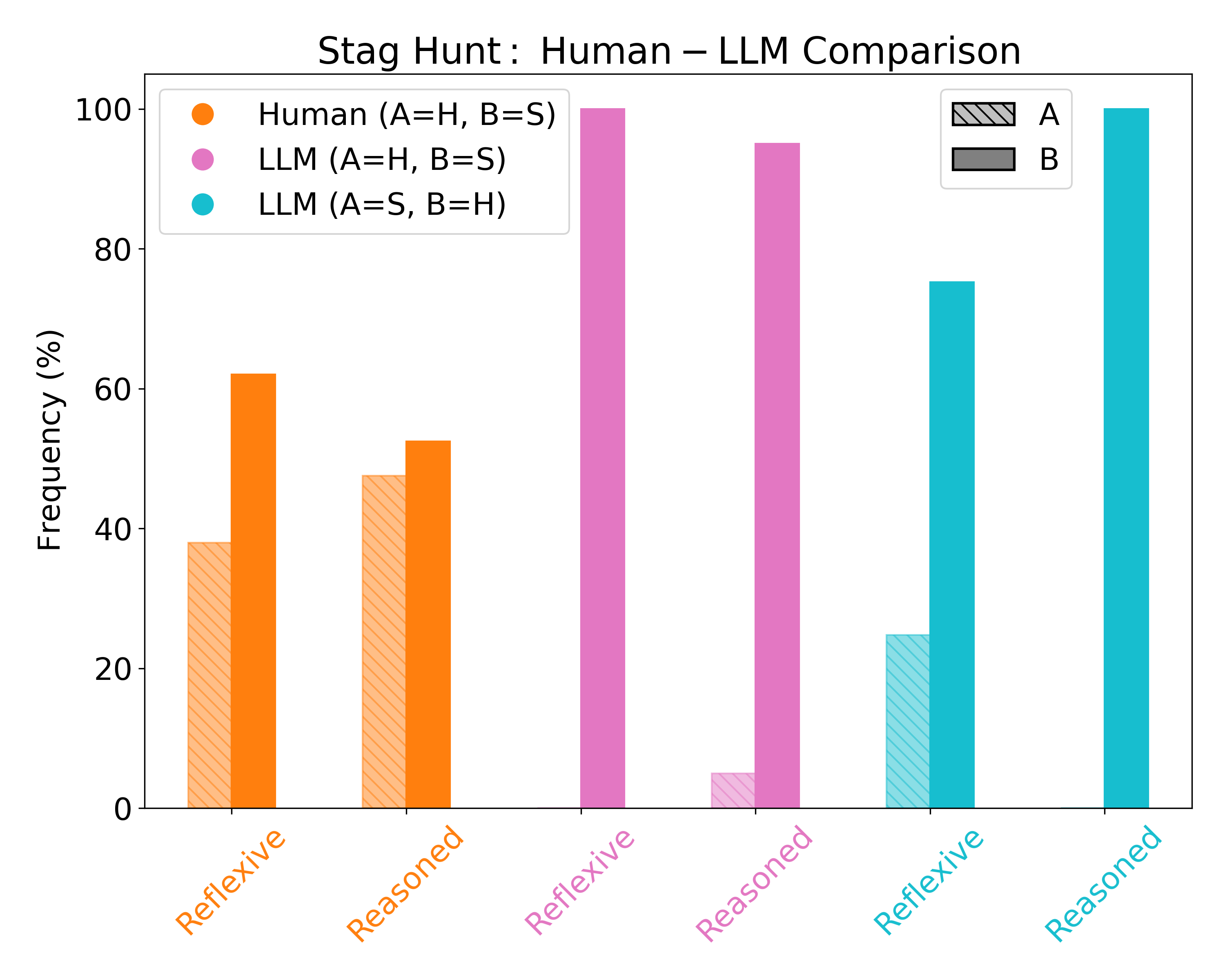}\hfill
\caption{Figure showing results for the preliminary study comparing the frequency at which Stag (S) and Hare (H) are selected by \textbf{{Humans}} where label A is Hare and B is Stag, an \textbf{{LLM}} where label A is Hare and B is Stag, and an \textbf{{LLM}} where label A is Stag and B is Hare. The LLM used in this preliminary study is GPT-4o. }
\label{fig:prelim}
\end{figure}

\citet{belloc2019intuition} present an experiment where a random sample of participants (88 people) play 4 games (352 games) of Stag Hunt under a time constraint (10 seconds), while a second set of participants (97) play 4 games (388 games) unconstrained. The constrained set of participants will be referred to as making a reflexive choice (without reasoning) and the unconstrained set make a reasoned choice. Processing the data made available by \citet{belloc2019intuition}, we found that when humans are making a reflexive choice, they typically select label B 62\% of the time and label A 38\% of the time. When making a reasoned choice, they still prefer label B, choosing it 52.5\% of the time, with label A being selected more often at 47.5\%. These results suggest that humans have a weak bias towards selecting label B, be it because of order, payoff, or something else entirely. These results can be seen in Figure \ref{fig:prelim}. 

We prompted (see HUMAN EQUIVALENT PROMPT below) GPT-4o with similar instructions and setting (where label A is Hare and B is Stag) the human participants were provided in \cite{belloc2019intuition} to compare the frequency at which label A and label B are selected. We also prompted GPT-4o under a different setting (one which the humans are not tested on), where label A is Stag and B is Hare, changing the meaning of the labels but keeping their order the same. We prompt GPT-4o to provide an answer only for the reflexive choice (400 games) and to reason over the task before providing the answer for the reasoned choice through CoT (400 games). We can see in Figure \ref{fig:prelim} that GPT-4o has a much stronger bias toward choosing label B than the human participants. Specifically, GPT-4o selects label B 97.5\% of the time in setting 1 and 87.65\% of the time in setting 2, on average. While human decision-making is based on both logical reasoning and potentially bias, GPT-4o is much more influenced by its biases. This suggests that there is considerable room for improvement in reducing the impact of these biases on the LLM's decision-making process.

\begin{tcolorbox}[breakable,colframe=black!75!black,title=HUMAN EQUIVALENT PROMPT]
In this experiment, you will be asked to make choices and you will have the opportunity to earn a certain amount of money. You will be paid privately at the end of the experimental session. Before you play each game you will be randomly and anonymously paired with another student of this lab section. The amount of money you will earn depends on your choice and on the choice of your partner. Your partner will not know your identity, just as you will not know his/hers. The interaction will only occur through the computer. You and your partner will simultaneously play the game. The matching mechanism will be such that you will play each new game with a new partner. In other words, you will never play twice with the same person. None of the people in this lab will know your choices.\\\\
You will be told the scores you can get playing the game and that will determine the amount of money you can earn at the end. The same information will be show to your partner. As we have already said, your score, and how much money you will earn, will depend on the choices done.\\\\
Scores you can get playing:\\\\
- If your choice will be {label\_1} and that of your partner will be {label\_1}, you will receive PO({label\_1}, {label\_1}) euros and your partner PO({label\_1}, {label\_1}) euro.\\\\
- If your choice will be {label\_1} and that of your partner will be {label\_2}, you will receive PO({label\_1}, {label\_2}) euros and your partner PO({label\_2}, {label\_1}) euro.\\\\
- If your choice will be {label\_2} and that of your partner will be {label\_1}, you will receive PO({label\_2}, {label\_1}) euros and your partner PO({label\_1}, {label\_2}) euro.\\\\
- If your choice will be {label\_2} and that of your partner will be {label\_2}, you will receive PO({label\_2}, {label\_2}) euros and your partner PO({label\_2}, {label\_2}) euro.
\label{prompt:human}
\end{tcolorbox}

\end{document}